\documentclass{article}

     \PassOptionsToPackage{numbers, compress}{natbib}
     \usepackage[final]{neurips_2022}

\usepackage[numbers]{natbib}

\usepackage[utf8]{inputenc} % allow utf-8 input
\usepackage[T1]{fontenc}    % use 8-bit T1 fonts
\usepackage{url}            % simple URL typesetting
\usepackage{booktabs}       % professional-quality tables
\usepackage{amsfonts}       % blackboard math symbols
\usepackage{nicefrac}       % compact symbols for 1/2, etc.
\usepackage{microtype}      % microtypography
\usepackage{xcolor}         % colors

\usepackage[colorlinks=true,linkcolor=blue,citecolor=blue]{hyperref}
\usepackage{amsmath,amsfonts,bm}

\def\Figref#1{Figure~\ref{#1}}

\def\secref#1{Section~\ref{#1}}
\def\Secref#1{Section~\ref{#1}}
\def\eqref#1{equation~\ref{#1}}
\def\1{\bm{1}}
\DeclareMathAlphabet{\mathsfit}{\encodingdefault}{\sfdefault}{m}{sl}
\SetMathAlphabet{\mathsfit}{bold}{\encodingdefault}{\sfdefault}{bx}{n}
\def\gN{{\mathcal{N}}}

\newcommand{\tabref}[1]{Table~\ref{#1}}
\usepackage{makecell}
\usepackage{color, colortbl}
\usepackage{wrapfig}
\usepackage{amsmath}
\usepackage{bbding}
\usepackage{pifont}
\usepackage{wasysym}
\usepackage{arydshln}
\usepackage{enumitem}

\usepackage{amsmath}
\usepackage{amssymb}
\usepackage{mathtools}
\usepackage{amsthm}

\usepackage[capitalize,noabbrev]{cleveref}

\theoremstyle{plain}

\theoremstyle{definition}

\theoremstyle{remark}

\usepackage[textsize=tiny]{todonotes}
\usepackage{multirow}

\definecolor{Gray}{gray}{0.9}

\newcommand{\nopretrain}{\textbf{No pretrain }}
\newcommand{\masktask}{\textbf{Node Prediction}}
\newcommand{\contexttask}{\textbf{Context Prediction }}
\newcommand{\motiftask}{\textbf{Motif Prediction }}
\newcommand{\contrastivetask}{\textbf{Contrastive learning }}
\newcommand{\supervisedmasktask}{\textbf{Masking Node + Supervised }}
\newcommand{\supervisedcontexttask}{\textbf{Context Prediction + Supervised }}
\newcommand{\supervisedtask}{\textbf{Supervised}}

\title{Does GNN Pretraining Help Molecular Representation?}

\author{
    Ruoxi Sun \\
 Google Cloud AI Research\\
  \texttt{ruoxis@google.com} \\
    \And
    Hanjun Dai \\
  Google Research, Brain Team\\
  \texttt{hadai@google.com} \\
  \And
  Adams Wei Yu\\
  Google Research, Brain Team\\
  \texttt{adamsyuwei@google.com} \\
}
\begin{document}

\maketitle

%\begin{abstract}
  
%\end{abstract}

\begin{abstract}
%\vspace{-.5em}
Extracting informative representations of molecules using Graph neural networks (GNNs) is crucial in AI-driven drug discovery. Recently, the graph research community has been trying to replicate the success of self-supervised pretraining in natural language processing, with several successes claimed. However, we find the benefit brought by self-supervised pretraining on small molecular data can be negligible in many cases. We conduct thorough ablation studies on the key components of GNN pretraining, including pretraining objectives, data splitting methods, input features, pretraining dataset scales, and GNN architectures, to see how they affect the accuracy of the downstream tasks. Our first important finding is, self-supervised graph pretraining do not always have statistically significant advantages over non-pretraining methods in many settings. Secondly, although noticeable improvement can be observed with additional supervised pretraining, the improvement may diminish with richer features or more balanced data splits. Thirdly, hyper-parameters could have larger impacts on accuracy of downstream tasks than the choice of pretraining tasks, especially when the scales of downstream tasks are small. Finally, we provide our conjectures where the complexity of some pretraining methods on small molecules might be insufficient, followed by empirical evidences on different pretraining datasets.

\end{abstract}
%\end{abstract}

\section{Introduction}

%The gain obtained by pretraining, mainly refer to supervised pretraining,  is larger in lower accuracy region (i.e. on scaffold split, basic feature) than that on high accuracy region (i.e. on balanced scaffold, rich feature). By using rich features, the gain achieved by supervised pretraining is reduced. 

%We reach a conclusion that unsupervised pretraining strategies do not have significance improvements over no pretraining, whereas supervised pretrainig improves more on low accuracy region, whereas the effect is reduced in higher accuracy region (i.e.achieved by balanced split or rich feature). 

Graph neural networks (GNNs)
, due to their effectiveness,
have been adopted to model a wide range of structured data, such as social networks, road graphs, citation networks, etc. Molecule modeling is one of these important applications, where it serves as the foundation of biomedicine and nurturing techniques like novel drug discovery. However, labeling biomedical data are usually time-consuming and expensive and thus task-specific labels are extremely inadequate. This poses a big challenge to the field. Recently, inspired by the remarkable success of self-supervised pretraining from natural language processing~\citep{devlin2018bert, brown2020language,wei2021finetuned} and computer vision domains~\citep{he2020moco,chen2020simple}, researchers start trying to apply the pretrain-finetune paradigm to molecule modeling with GNN, hoping to boost the performance of various molecular tasks by pretraining the model on the enormous unlabeled data.
For instance, %a large variety of 
many methods have been proposed~\citep{xie2021self,wu2021self}, where significant performance improvements are claimed
% due to well designed pretraining objectives and 
by pretraining on large scale datasets \citep{hu2019strategies, rong2020self,  you2020graph, you2020does, zhu2021graph}. Despite of the promising results, %we find research on investigating the factors for the improvement are still at a nascent stage. Based on our experiments, 
we find that reproducing some of these outstanding gains via graph pretraining can be non-trivial, and sometimes the improvement largely relies on the experimental setup and the extensive hyper-parameter tuning of downstream tasks, rather than the design of pretraining objectives. These observations motivate us to rethink the effectiveness of graph pretraining with unsupervised or self-supervised objectives, and investigate what factors would influence the effectiveness of self-supervised graph pretraining.
% and to investigate what possibly influences the effectiveness. % All of these motivate this paper.

\begin{figure*}[ht]
\includegraphics[width=1.0\textwidth]{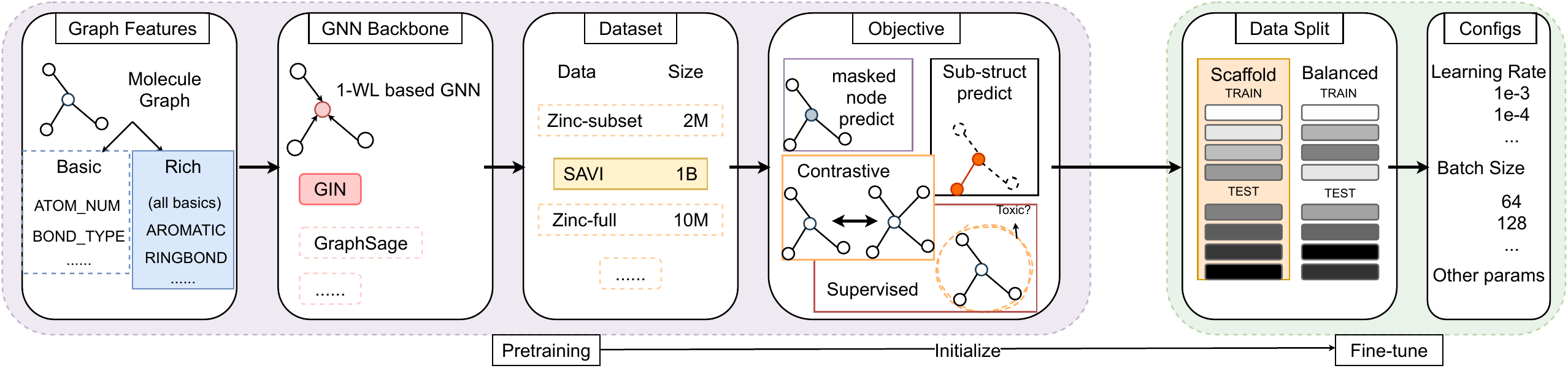}
\caption{A typical pipeline for graph pretraining and deployment for downstream applications. \label{fig:pipeline}}
\end{figure*}

In this paper, we perform systematic studies to
assess the performance of popular graph pretraining objectives on different types of datasets, and exploit various confounding components in experimental setup in deciding the performance of downstream tasks with or without pretraining. 
Here, we restrict our studies to small molecular graphs, as opposed to other application domains, such as social networks or citation graphs. The key insights and take-aways of this paper are:
\begin{itemize}[leftmargin=*]%[noitemsep,topsep=1pt,parsep=1pt,partopsep=0pt, leftmargin=*]
    \item Among the pretraining tasks we evaluated, the self-supervised pretraining alone does not provide statistically significant improvements over non-pretrained methods on downstream tasks.
    \item When additional supervised pretraining step is conducted after self-supervised pretraining, we observe statistically significant improvements. However, the gain becomes marginal on some specific data splits or diminishes if richer features are introduced.
    \item Beyond data splits and hand-crafted features, the usefulness of graph pretraining is also sensitive to the experimental hyperparameters, such as learning rates and number of study repeats. Different setups can lead to opposite conclusions. 
    \item In conclusion, different from the previous works, we do not observe clear and unconditional gains achieved by graph pretraining on molecular representation, indicating it is still too early to conclude graph pretraining is effective in molecular domain.
    \item  We investigate the reason of above and hypothesize that the complexity of some pretraining methods on molecules is insufficient, leading to less transferable knowledge for downstream tasks.
\end{itemize}

Despite the overall negative results we obtained, the main goal of this paper is not to discourage the pretraining research for small molecules. Instead, we hope to raise the attention on different aspects of experiments and the role of simple hand-crafted features, so as to provide useful information for designing better pretraining approaches. Below we first introduce the background of GNN and its pretraining in \secref{sec:gnn_pretrain}, and then our experimental design and results in \secref{sec:exp_framework} and \secref{sec:results}, respectively. Finally we conclude with our findings and the limitations in \secref{sec:summary} and  \secref{sec:limitation}.

\section{Preliminary}
\label{sec:gnn_pretrain}

\begin{table*}[th]
\centering
%\vspace{-5mm}
\caption{Summary of Experiments. \tabref{tab:graphsage + neurips split} and \tabref{tab:graphsage + iclr split} are deferred to appendix due to space limit. \label{tab:summary-table}}
\resizebox{0.9\textwidth}{!}{%
\begin{tabular}{c|cc|cc|cc|cc|cc}
%\hline
\Xhline{2\arrayrulewidth}
 &
  \multicolumn{2}{c|}{\textbf{Pretrain Objective}} &
  \multicolumn{2}{c|}{\textbf{Graph Features}} &
  \multicolumn{2}{c|}{\textbf{Downstream Splits}} &
  \multicolumn{2}{c|}{\textbf{GNN Arch}} &
  \multicolumn{2}{c}{\textbf{Pretrain Dataset}} \\ 
  \cline{2-11}
  %\xchline{2-11}{2\arrayrulewidth}
 &
{Self-Supervised} &
{Supervised} &
{Rich} &
{Basic} &
{Balanced} &
{Scaffold} &
{GIN} &
{GraphSage} &
{ZINC15} &
{SAVI} \\
  %\hline
  \Xhline{2\arrayrulewidth}
\textbf{\tabref{tab:unsupervised-obj}} 
   &\ding{52}
   &
   &\ding{52}
   &
   &\ding{52}
   &
   &\ding{52}
   &
   &\ding{52}
   &
   \\\hdashline
\textbf{\tabref{tab:super-obj}} 
   &
   &\ding{52}
   &\ding{52}
   &
   &\ding{52}
   &
   &\ding{52}
   &
   &\ding{52}
   &
   \\\hdashline
\textbf{\tabref{tab:split}} 
   &\ding{52}
   &
   &\ding{52}
   &
   &
   &\ding{52}
   &\ding{52}
   &
   &\ding{52}
   &
   \\\hdashline
\textbf{\tabref{tab:feature-scaffold-richfeature-supervised}} 
   &
   &\ding{52}
   &\ding{52}
   &
   &
   &\ding{52}
   &\ding{52}
   &
   &\ding{52}
   &
   \\\hdashline
\textbf{\tabref{tab:Balanced scaffold + basic feature + unsupervised}} 
   &\ding{52}
   &
   &
   &\ding{52}
   &\ding{52}
   &
   &\ding{52}
   &
   &\ding{52}
   &
   \\\hdashline
\textbf{\tabref{tab:Balanced scaffold + basic feature + supervised}}
   &
   &\ding{52}
   &
   &\ding{52}
   &\ding{52}
   &
   &\ding{52}
   &
   &\ding{52}
   &
   \\\hdashline
\textbf{\tabref{tab:Scaffold + basic feature + unsupervised} }
   &\ding{52}
   &
   &
   &\ding{52}
   &
   &\ding{52}
   &\ding{52}
   &
   &\ding{52}
   &
   \\\hdashline
\textbf{\tabref{tab:Scaffold + basic feature + supervised}}
   &
   &\ding{52}
   &
   &\ding{52}
   &
   &\ding{52}
   &\ding{52}
   &
   &\ding{52}
   &
   \\\hdashline
%\textbf{\tabref{tab:graphsage + neurips split}} 
%   &\ding{52}
%   &\ding{52}
%   &\ding{52}
%   &
%   &\ding{52}
%   &
%   &
%   &\ding{52}
%   &\ding{52}
%   &
%   \\ \hdashline
%\textbf{\tabref{tab:graphsage + iclr split}} 
%   &\ding{52}
%   &\ding{52}
%   &\ding{52}
%   &
%   &
%   &\ding{52}
%   &
%   &\ding{52}
%   &\ding{52}
%   &
%   \\\hdashline
\textbf{\tabref{tab:savi-balanced-split}} 
   &\ding{52}
   &\ding{52}
   &\ding{52}
   &
   &\ding{52}
   &
   &\ding{52}
   & 
   &
   &\ding{52}
   \\\hdashline
\textbf{\tabref{tab:SAVI-scaffold-split}} 
   &\ding{52}
   &\ding{52}
   &\ding{52}
   &
   &
   &\ding{52}
   &\ding{52}
   &
   &
   &\ding{52}
   \\ 
   %\hline
  % \Xhline{2\arrayrulewidth}
  \hdashline 
   \textbf{\tabref{tab:graphsage + neurips split}} 
   &\ding{52}
   &\ding{52}
   &\ding{52}
   &
   &\ding{52}
   &
   &
   &\ding{52}
   &\ding{52}
   &
   \\ \hdashline
\textbf{\tabref{tab:graphsage + iclr split}} 
   &\ding{52}
   &\ding{52}
   &\ding{52}
   &
   &
   &\ding{52}
   &
   &\ding{52}
   &\ding{52}
   &
   \\\hdashline
     \Xhline{2\arrayrulewidth}
\end{tabular}%
%\vspace{-5mm}
}
\end{table*}

%\section{Graph Pretraining}
%Some data like chemical molecules contains structure information, which can be modeled as a graph. For example, in chemical molecules, atoms are modeled as nodes, and bonds are modeled as edge. Start from some format of node and edge feature, Graph neural networks performs several rounds of convolution of every node using the features of its connected neighbour nodes and the edges between them. After several sounds of convolution, the node feature and edge feature are Therefore, it is a natural way to represent raw format of data into deep-learning usable representation while maintain structure information.  THe representation can be used on pretraining tasks or downstream tasks with different objectives. 
% {\color{red} In graph learning, there
% are usually two categories of supervised tasks: i) Node classification/regression, where each node
% v has a label/target yv, and the task is to learn to predict the labels of unseen nodes; ii) Graph
% classification/regression, where a set of graphs {G1, ..., GN } and their labels/targets {y1, ..., yN } are
% given, and the task is to predict the label/target of a new graph.

% %we divide existing graph SSL methods into three categories: contrastive, generative, and predictive,

% }

%We choose implementation of ICLR paper \citet{hu2019strategies}, as we can reproduce the results reported in the paper, and we can observe gains on the proposed scaffold split under the authors setup. 

\paragraph{Graph Neural Networks (GNNs).} Let $G = \{V, E\}$ denote a molecule graph with $V$ as the set of nodes and $E$ as the set of edges. 
%$X_v$ as node attributes for $v \in V$ and $e_{ij}$ as edge features for edge(i, j) $\in E$. 
Given the node features $X_i$, most GNNs learn an embedding representation $h_i$ for every node $i \in V$ by aggregating representations from connected nodes and edges, denoted as graph convolution. These procedure repeats for $K$ times with the update equation as follows:
\begin{equation}
%\resizebox{1.0\hsize}{!}{$
h_i^k = \text{UPDATE}(h_i^{k-1}, \text{AGGREGATE}(\{h_i^{k-1}, h_j^{k-1}, e_{ij}\}: \forall j \in N(i))) \nonumber
%$}
\end{equation}
where $\gN(i)$ is the set of neighbor nodes of $i$ and $h_i^0 = X_i$. The representation for entire graph $G$ is then obtained by permutation-invariant transformation on node representation,  $h_G = \text{READOUT}(h_i^K | i\in V)$. In this paper we mainly study the GNNs that belong to this family, namely the WL-1 GNNs.

\paragraph{Finetune.} After the pretraining, the pretrained model is used to finetune on the downstream tasks. For molecule property prediction tasks, the graph-level representation obtained from the pretrained model is connected to linear classifiers to predict downstream task labels. The fine-tuning is performed in an end-to-end manner, where both the pretrained GNN and the linear classifiers are trainable. %: $y_G = f_F(h_G)$, where $h_G = \text{GNN}_{\theta}(X_G)$. the notation is terrible, so I'll just delete these for simplicity.

\paragraph{Graph pretraining objectives.} The primary goal of pretraining is to learn representations with robust transferable knowledge of graphs from the abundant pretraining data %through well-designed pretraining objectives. The learned knowledge can be 
and then generalize to downstream tasks with usually different supervision signals. Generally the pretraining objectives can be categorized into self-supervised and 
supervised ones. We present a brief overview of some representative objectives in the following sections. 

\subsection{Self-supervised (unsupervised) pretraining}
In self-supervised pretraining, the pretraining objective is designed to learn self-generated targets from the structure of the molecules, such as the type of nodes and edges, prediction of local context, graph partition, node clustering, occurrence of some functional groups, and etc. The predictive target can be node/edge level or entire graph level. We present some representative ones below:

{\color{red} %e both node and graph embeddings are of high-quality so that graph embeddings are robust and transferable across downstream tasks

%Therefore, while this type of pre-training is also very natural, it is beyond the scope of this paper and we leave its investigation for future work.

% When the GNN pre-training is finished, we fine-tune the pre-trained GNN model on downstream tasks. Specifically, we add linear classifiers on top of graph-level representations to predict downstream graph labels. The full model, i.e., the pre-trained GNN and downstream linear classifiers, is subsequently fine-tuned in an end-to-end manner. 

}
\subsubsection{Node Prediction}
Node prediction is a node-level classification task given the masked context of entire graph. Similar to \citet{devlin2018bert}, some portion of node attributes are masked and replaced with mask-specified indicators in the node input feature. After graph convolution, %aggregating neighbor representations, 
the embedding output from GNN is used to predict the true attribute of the node, e.g. atom type in molecular graphs, through a linear classifier on top of the node embedding. 
%The training objective is % $\min \sum_i \text{loss}(f_N(h_i), T_i)$ where $GNN$
%the   The node attribute is masked and the pretraining GNN is to predict the type of node (atom) 
%\cite{devlin2018bert}
%{\color{red} In Attribute Masking, we aim to capture domain knowledge by learning the regularities of the node/edge attributes distributed over graph structure.  Attribute Masking pre-training works as follows: We mask node/edge attributes and then we let GNNs predict those attributes (Devlin et al., 2019) based on neighboring structure.  Specifically, We randomly mask input node/edge attributes, for example atom types in molecular graphs, by replacing them with special masked indicators. We then apply GNNs to obtain the corresponding node/edge embeddings (edge embeddings can be obtained as a sum of node embeddings of the edge’s end nodes). Finally, a linear model is applied on top of embeddings to predict a masked node/edge attribute. }
%\begin{equation}
%   \min_{\theta, \gamma} \sum_i \text{Loss}(f_{N_\gamma}(h_i), T_i ) \quad \text{where } h_i = \text{GNN}_{\theta}(X_i)  
% \end{equation}

\subsubsection{Context Prediction}
Context prediction task is a sub-graph level task aiming at learning embedding that can represent the local subgraph surrounding a node. Generally it can be viewed as a masked task for substructure. Since it is essentially a structured prediction which can be difficult in general, \citet{hu2019strategies} leverages the adversarial learning to teach the model to distinguish the positive sub-graph embedding from the negative ones. 
%dot product of the two embedding. 
\citet{rong2020self} instead builds a dictionary of structures that captures the property of sub-graphs (e.g. type and quantity of neighbour nodes and bonds), and turns it into a multi-class classification problem. %using the embedding of node in the context center. %$h_i^K$. 
% {\color{red} In Context Prediction, we use subgraphs to predict their surrounding graph structures. Our goal
% is to pre-train a GNN so that it maps nodes appearing in similar structural contexts to nearby embeddings. 
% \textbf{GROVER} For the node/edge-level
% tasks, instead of predicting the node/edge type alone, GROVER randomly masks a local subgraph
% of the target node/edge and predicts this contextual property from node embeddings.
% The prediction target should reflect contextual information of the node/edge. Guided by these
% criteria, we present the tasks on both nodes and edges.
% They both try to predict the context-aware properties of
% the target node/edge within some local subgraph. 
% }
%\begin{equation}
%    % \quad
%    \min_{\theta} \sum_i \text{loss}(C_i, h_i) \quad \text{where } 
%    \quad
%    C_i = f_C(h_i, h_j: j\in \text{context}(i))
%\end{equation}
%where $f_c$ performs some transformation on context and target node to generate context embedding or labels. $f_C$ is embedding extraction function for \cite{hu2019strategies} and is label extraction for \cite{rong2020self}.

\subsubsection{Motif Prediction}
Motif prediction~\citep{rong2020self} is to %learn graph-level embedding through 
predict the existence of functional groups, such as benzene ring or hydroxyl. The motifs are extracted automatically from RDKit~\citep{landrum2016rdkit} . The motif prediction task is formulated as a graph-level multi-label binary classification task, where the graph embedding is used to jointly predict the occurrence of these semantic functional motifs.  %with one label for one motif. 
%Given a molecule, computational sorftware  can provide the supervised labels, and one label for one motif. Hence motif prediction task is a self-supervised task. The training objective is   
%
%\begin{equation}
%   \min_{\theta, \omega}  \text{Loss}(f_{M_\omega}(h_G), \{M_j: j = 1,2, .. m \} ) \quad \text{where } h_G = \text{GNN}_{\theta}(X) 
% \end{equation}
%where  $\{M_j: j = 1,2, .. m \}$ is a set of binary labels for $m$ motifs. $f_{M_\omega}$ is a multi-task linear classifier parameterized by $\omega$. 

%{\color{red} For the graph-level tasks, by incorporating the domain knowledge, GROVER extracts
%the semantic motifs existing in molecular graphs and predicts the occurrence of these motifs for a molecule from graph embeddings.

%fs are recurrent sub-graphs among the input graph data, which are prevalent in molecular graph data. One important class of motifs in molecules are functional groups, which encodes the rich domain knowledge of molecules and can be easily detected by the professional software, such as RDKit [27]. Formally, the motif prediction task can be formulated as a multi-label classificationproblem, where each motif corresponds to one label. }
\subsubsection{Contrastive learning}
Graph contrastive learning is to %, pretraining is performed through
maximize the agreement of two augmented views of the same graph, and minimize the agreement of different graphs. The optimization is conducted using contrastive loss in the latent embedding space \citep{you2020graph, hassani2020contrastive,zhu2021graph,wu2021self}. The augmentation function needs to transform graphs into realistic and novel augmentations without affecting semantic labels of the graphs. For example, the transformation can be small perturbations or modifications on node/edge embedding, drop of a few nodes or edges, and so on. These transformations enforce an underlying prior for contrastive learning, that is, local transformation does not change the semantic meaning of a graph.
\subsection{Supervised pretraining}
%{\color{red} We inject graph-level domain-specific knowledge into our pretrained embeddings by defining supervised graph-level prediction tasks In particular, we consider a practical method to pre-train graph representations: graph-level multi-task supervised pre-training to jointly predict a diverse set of supervised labels of individual graphs. }
Supervised pretraining aims to learn domain-specific graph-level knowledge from specifically designed pretraining tasks. For molecular application, the supervised labels are generated from a diverse set of functional studies like biochemical assays. The pretrainning task is to perform multiple binary classification and jointly learn the supervised labels. Although the pretraining mainly refers to unsupervised or self-supervised methods as they are not limited by the requirement of supervised labels, supervised pretraining is still a great source to investigate the graph pretraining in general.

\begin{table*}[t]
\small
\centering
%\vspace{-2mm}
\caption{\textbf{Self-supervised} + Rich feature + Balanced Scaffold Split.  No pretrain has an average value of 78.0\% over all 5 datasets. \label{tab:unsupervised-obj}}
\vspace{2mm}
\resizebox{\textwidth}{!}{%
\begin{tabular}{lcccccr}
\Xhline{2\arrayrulewidth}
%\hline
Methods           & \textbf{BBBP}            & \textbf{BACE}                  & \textbf{TOX21}           & \textbf{TOXCAST}        & \textbf{SIDER  }  &    \textbf{AVE GAIN}    \\ %\midrule %\hline
\Xhline{2\arrayrulewidth}
\nopretrain      & 92.23($\pm$3.07) & 87.43($\pm$1.63)     & 79.20($\pm$1.99)  & 69.13($\pm$0.55) & 61.92($\pm$0.89) & 0($\pm$1.626)\\
\masktask        & 92.24($\pm$2.76) & 87.32($\pm$1.67)     & 79.57($\pm$2.03) & 69.77($\pm$0.13) & 61.62($\pm$1.12) & 0.122($\pm$1.542)\\
\contexttask     & 92.68($\pm$1.19) & 86.98($\pm$1.26)     & 79.05($\pm$2.51) & 70.18($\pm$0.44) & 61.65($\pm$0.77) & 0.126($\pm$1.234)\\
\motiftask       & 92.63($\pm$1.19) & 87.16($\pm$1.66)     & 79.22($\pm$2.38) & 69.09($\pm$0.07) & 62.45($\pm$1.25) & 0.128($\pm$1.310)\\  
\contrastivetask  & 92.31($\pm$1.58)& 86.67($\pm$2.40)     & 78.45($\pm$2.44) & 68.37($\pm$0.80) & 61.22($\pm$1.20) & -0.578($\pm$1.684) \\ 
%\hline
\Xhline{2\arrayrulewidth}
%\vspace{-6mm}
\end{tabular}%
}
\vspace{-2mm}
\end{table*}

\begin{table*}[h]
\small
%\caption{Supervised Pretraining Objectives on Downstream Task Accuracy (\%)}
%\vspace{-2mm}
\centering
\caption{\textbf{Supervised} + Rich feature + Balanced Scaffold.  No pretrain has an average AUC of 78.0\%. \label{tab:super-obj}}
\vspace{2mm}
\resizebox{\textwidth}{!}{%
\begin{tabular}{lcccccr}
%\hline
\Xhline{2\arrayrulewidth}
Methods                    & \textbf{BBBP}   & \textbf{BACE}   & \textbf{TOX21}   & \textbf{TOXCAST} & \textbf{SIDER}  &    \textbf{AVE GAIN}   \\ 
\Xhline{2\arrayrulewidth}
\nopretrain      & 92.23($\pm$3.07) & 87.43($\pm$1.63) & 79.2($\pm$1.99)  & 69.13($\pm$0.55) & 61.92($\pm$0.89) & 0($\pm$1.626)\\
\supervisedtask  & 91.65($\pm$2.11) & 86.91($\pm$1.86) & 81.13($\pm$2.39) & 71.64($\pm$0.46) & 62.14($\pm$1.13) & 0.712($\pm$1.590)\\
\supervisedmasktask     & \textbf{93.43($\pm$2.50)} & 86.90($\pm$2.04) & \textbf{81.93($\pm$1.79)} & 71.66($\pm$0.73) & 62.68($\pm$1.82) & 1.338($\pm$1.776)\\
\supervisedcontexttask& 92.27($\pm$1.57) & \textbf{88.72($\pm$1.68)} & 81.71($\pm$1.79)   & \textbf{72.19($\pm$0.79)}  & \textbf{63.21($\pm$1.49)}  & 1.638($\pm$1.464) \\ 
\Xhline{2\arrayrulewidth}
%\vspace{-10mm}
\end{tabular}%
}
\vspace{-2mm}
\end{table*}

\begin{table*}[t]
\small
%\vspace{-3mm}
\centering
\caption{Self-supervised + Rich feature + \textbf{Scaffold}. No pretrain has an average ROC-AUC of 71.8\% over all benckmark datasets. \label{tab:split}}
\vspace{2mm}
\resizebox{\textwidth}{!}{%
\begin{tabular}{lcccccr}
%\hline
\Xhline{2\arrayrulewidth}
Methods        & \textbf{BBBP}             & \textbf{BACE}    & \textbf{TOX21}    & \textbf{TOXCAST} & \textbf{SIDER} &    \textbf{AVE GAIN}   \\ %\hline
\Xhline{2\arrayrulewidth}
%\nopretrain      & \textbf{75.13($\pm$1.08)} & 79.79($\pm$1.18) &  76.06($\pm$0.20) & 66.1($\pm$0.42)  & 61.21($\pm$0.76) \\
\nopretrain      & 74.83($\pm$0.73) & 80.10($\pm$0.42) & 75.86($\pm$0.58)  & 65.95($\pm$0.15) & 62.30($\pm$1.14) & 0($\pm$0.579)\\
\masktask        & 73.45($\pm$0.27) & 83.66($\pm$0.75) & 75.30($\pm$0.37)  & 66.50($\pm$0.06) & 65.08($\pm$0.12) & 0.990($\pm$0.323)\\
\contexttask     & 74.10($\pm$0.22) & 81.87($\pm$0.49) &75.37($\pm$0.11)   & 66.86($\pm$0.07) & 62.84($\pm$0.46) & 0.400($\pm$0.280)\\
\motiftask       & 73.65($\pm$0.36) & 80.58($\pm$2.04) & 74.55($\pm$0.79)  & 65.63($\pm$0.07) & 64.05($\pm$0.23) & -0.116($\pm$0.766)\\
\contrastivetask & 73.32($\pm$2.38) & 80.51($\pm$0.80) & 74.55($\pm$0.22)  & 65.70($\pm$0.09) & 64.39($\pm$0.63) & -0.114($\pm$0.513)\\ 
%\hline
\Xhline{2\arrayrulewidth}
%\vspace{-6mm}
\end{tabular}%

}
%\caption{Scaffold split + rich feature}
\end{table*}

\begin{table*}[t]
\small
\centering
%\vspace{-3mm}
\caption{Supervised %+ Self-supervised 
+ Rich feature + \textbf{Scaffold}. No pretrain get 71.8\% average ROC-AUC. \label{tab:feature-scaffold-richfeature-supervised}}
\vspace{2mm}
\resizebox{\textwidth}{!}{%
\begin{tabular}{lcccccr}
%\hline
\Xhline{2\arrayrulewidth}
Methods & \textbf{BBBP}              & \textbf{BACE}     & \textbf{TOX21}    & \textbf{TOXCAST}      & \textbf{SIDER} &    \textbf{AVE GAIN}  \\ %\hline
\Xhline{2\arrayrulewidth}
\nopretrain                 & \textbf{74.83($\pm$0.73)}     & 80.10($\pm$0.42)          & 75.86($\pm$0.58)          & 65.95($\pm$0.15)          & 62.30($\pm$1.14)           & 0($\pm$0.604) \\
\supervisedtask             & 72.79( $\pm$ 0.7)             & 83.23($\pm$0.67)          & 77.66($\pm$0.08)          & 67.72($\pm$0.13)          & 65.34($\pm$0.17)           & 1.540($\pm$0.350) \\
\supervisedmasktask         & 73.38($\pm$0.55)              & \textbf{84.42($\pm$0.27)} & \textbf{77.85($\pm$0.24)} & \textbf{67.14($\pm$0.28)} & 64.06($\pm$0.28)           & 1.562($\pm$0.324) \\
\supervisedcontexttask      & 73.81($\pm$0.52)              & \ 84.35($\pm$0.93)        & 77.11($\pm$0.14)          & 67.87($\pm$0.08)          & \textbf{65.19($\pm$0.17)}  & 1.858($\pm$0.368)  \\ 
%\hline
\Xhline{2\arrayrulewidth}
%\vspace{-7mm}
\end{tabular}%
}
\end{table*}

\section{Experiment framework}
\label{sec:exp_framework}
%\vspace{-3mm}
%To design our experimental framework 
%To understand more about the 
To investigate pretraining on graphs for molecule representations, we first revisit the typical pretraining-finetuning pipeline used in the literature. \Figref{fig:pipeline} shows the overall procedure of deployment, with several design choices presented at each stage of the pipeline. Since different choices at each stage can lead to different performances on the downstream tasks, 
we investigate them one at a time while keeping others the unchanged. %
The design principle of our experiment framework is to analyze the effect of \emph{every} stage in the pipeline as comprehensive as possible, while also keeping it tractable to avoid exponentially many experiments. %Below we describe these stages and the corresponding design choices we considered in the experiments. 
%\vspace{-3mm}
\subsection{Design choices}
\label{sec:design_choice}
%\vspace{-4mm}
We consider the design choices for the four pretraining objectives.
%\vspace{-1mm}
\paragraph{Pretraining objective} In \Secref{sec:gnn_pretrain} we have provided a brief literature review over the pretraining methods for molecule representation. Here we categorize those pretraining by different principles, and present one well-recognized representative of each category. The representatives are selected because they have more desired properties, such as better performance, compared with their counterparts. 
\begin{itemize}[noitemsep,topsep=3pt,parsep=3pt,partopsep=5pt, leftmargin=*]
	\item \textbf{Masking.} We leverage the node prediction objective, which randomly masks 15\% of the nodes' feature and then ask GNN to make prediction on the node attributes of the masked ones. This strategy %is %from~\citet{hu2019strategies} and 
	resembles the BERT pretraining~\citep{devlin2018bert} in natural language processing.
	\item \textbf{Structured.} Unlike text data where the topology is a sequence, the graph has rich structure information. Following~\citet{hu2019strategies}, we use context prediction objective, which masks out the context from $k_1$-hops to $k_2$-hops and leverages  adversarial training to predict the true context embeddings from the random context embeddings. 
	\item \textbf{Graph-level self-supervised.} Following \citep{rong2020self}, GNN is asked to predict whether a motif is contained in a molecule. The motif can be extracted from the molecule with RDKit~\citep{landrum2016rdkit}. %We denote it as \textbf{Motif Prediction}, and use the same set 
The motifs are 85 motifs~\footnote{\url{http://rdkit.org/docs/source/rdkit.Chem.Fragments.html}} %as~\citep{rong2020self}
	for multi-label classification. 
	\item \textbf{Contrastive.}  We generate two views of the same graph by corrupting the input node features with Gaussian noise. %to generate different views of the same graph, %and leverage the locally normalized probability to
	We leverage the contrastive learning loss proposed in ~\citep{you2020graph}: we maximize the consistency between positive pairs (from same graphs) and minimize that between negative pairs (from different graphs). In this paper, we restrict ourselves to this specific contrastive training method, however, various contrastive learning methods can be further explored. 
	\item \textbf{Graph-level supervised.} Finally when applicable, we use the ChEMBL dataset with graph-level labels for graph-level supervised pretraining as ~\citet{hu2019strategies}. 
\end{itemize}

The above are the design choices for pretraining objectives. %In addition to these stages, we believe the factors in downstream applications can potentially affect the effectiveness of pretraining as well. 
Next, we consider other factors that influence graph-pretraining performance. %the configurations for the finetuning stages in the downstream tasks. 
%\vspace{-4mm}
%\paragraph{Learning hyperparameters for downstream tasks} We find that the models initialized from pretrained checkpoints and those from scratch may favor different hyperparameter configurations like learning rates. So to make a fair comparison, we would tune the learning rate separately for each setting, instead of using the same configuration for all the settings. This is critical for the fair comparison of performance on downstream tasks. 
%\vspace{-4mm}
\paragraph{Graph Features} Each molecule is represented by a graph with atoms as nodes and bonds as edges. In this paper we mainly consider the graph representations without the 3D information. %Generally speaking, in each of such a graph, each node (atom) or edge (bond) is associated with a binary feature vector. The features are constructed based on domain knowledge, and 
%The features prepared for graph neural networks are 
For each molecule graph, chemical properties of nodes and edges are extracted to serve as node and edge features for the graph neural networks. Depending on how rich the features are, we categorize the design choices into two categories:
\begin{itemize}[noitemsep,topsep=1pt,parsep=1pt,partopsep=0pt, leftmargin=*]
	\item \textbf{Basic features.} The basic set of features are the ones used in \citet{hu2019strategies}. Specifically, the node features contain the atom type and the derived features, such as formal charge list, chirality list, etc. The edge features contain the bond types and the bond directions. These features are categorical, and thus will be encoded in a one-hot vector individually and then concatenated together to form the feature vector for node/edge representation. 
	\item \textbf{Rich features.} The rich feature set is a superset of the basic features. In addition to the basic ones mentioned above, it comes with the additional node features such as hydrogen acceptor match, acidic match and bond features such as ring information. This set of features are used in ~\citet{rong2020self}. Additionally and importantly, we follow their setting to incorporate additional 2d normalized rdNormalizedDescriptors features~\footnote{ \url{https://github.com/bp-kelley/descriptastorus} for the feature descriptor.}, which is used in the downstream tasks only and not in pretraining.
\end{itemize}
Please refer to the original papers for the full set of basic~\citep{hu2019strategies} and rich~\citep{rong2020self} features, respectively.

\begin{table*}[t]
\small
%\vspace{-3mm}
\centering
\caption{Self-supervised + \textbf{Basic feature} + Balanced Scaffold. No pretrain has an average AUC of 76.7\% over all 5 datasets. \label{tab:Balanced scaffold + basic feature + unsupervised}}
\vspace{2mm}
\resizebox{\textwidth}{!}{%
\begin{tabular}{lcccccr}
%\hline
\Xhline{2\arrayrulewidth}
Methods                   & \textbf{BBBP}                           & \textbf{BACE}                           & \textbf{TOX21}                          & \textbf{TOXCAST}                        & \textbf{SIDER}        &    \textbf{AVE GAIN}                    \\ 
%\hline
\Xhline{2\arrayrulewidth}
\nopretrain              & 91.46($\pm$ 0.85) &  84.29($\pm$ 3.80) &  78.35($\pm$ 0.95)  & 68.31($\pm$ 1.61) & 61.15($\pm$ 2.46)  &  0($\pm$1.934)\\
\masktask                & 91.23($\pm$ 1.51) & 84.97($\pm$ 1.55) & 77.77($\pm$ 1.23) & 68.98($\pm$ 1.11) & 61.20($\pm$ 0.41)  &   0.118($\pm$1.162)\\
\contexttask             & 92.13($\pm$ 1.04) & 84.83($\pm$ 3.19) & 78.79($\pm$ 2.52) & 68.29($\pm$ 1.23) & 62.32($\pm$ 2.99) &   0.560($\pm$2.194)\\
%\supervisedtask              & 90.70($\pm$ 0.74)  & 84.22($\pm$ 2.69) & 80.45($\pm$ 1.47) & 69.47($\pm$ 1.06) & 63.38($\pm$ 1.44) \\
%\supervisedmasktask     & 91.10($\pm$ 2.88)  & 85.54($\pm$ 4.57) & 81.49($\pm$ 1.52) & 70.77($\pm$ 1.00)  & 62.81($\pm$ 2.61) \\
%\supervisedcontexttask & 91.54($\pm$ 3.52) & 85.71($\pm$ 2.92) & 81.23($\pm$ 1.94) & 71.36($\pm$ 1.05) & 62.75($\pm$ 2.27) \\ 
%\hline
\Xhline{2\arrayrulewidth}
\end{tabular}
}
\end{table*}

\begin{table*}[t]
% \small
%\vspace{-4mm}
\centering
\caption{Supervised + %Self-supervised +
\textbf{Basic feature} + Balanced Scaffold. No pretrain has an average AUC of 76.7\%. \label{tab:Balanced scaffold + basic feature + supervised}}
%\vspace{-3mm}
\resizebox{\textwidth}{!}{%
\begin{tabular}{lcccccr}
%\hline
\Xhline{2\arrayrulewidth}
Methods                    & \textbf{BBBP}                           & \textbf{BACE}                           & \textbf{TOX21}                          & \textbf{TOXCAST}                        & \textbf{SIDER}     &    \textbf{AVE GAIN}                       \\ 
%\hline
\Xhline{2\arrayrulewidth}
\nopretrain            & 91.46($\pm$ 0.85) &  84.29($\pm$ 3.80) &  78.35($\pm$ 0.95)  & 68.31($\pm$ 1.61) & 61.15($\pm$ 2.46) & 0($\pm$ 1.934) \\
%\masktask                & 91.23($\pm$ 1.51) & 84.97($\pm$ 1.55) & 77.77($\pm$ 1.23) & 68.98($\pm$ 1.11) & 61.20($\pm$ 0.41)  \\
%\contexttask             & 92.13($\pm$ 1.04) & 84.83($\pm$ 3.19) & 78.79($\pm$ 2.52) & 68.29($\pm$ 1.23) & 62.32($\pm$ 2.99) \\
\supervisedtask              & 90.70($\pm$ 0.74)  & 84.22($\pm$ 2.69) & 80.45($\pm$ 1.47) & 69.47($\pm$ 1.06) & \textbf{63.38($\pm$ 1.44)} & 0.932($\pm$ 1.480)\\
\supervisedmasktask     & 91.10($\pm$ 2.88)  & 85.54($\pm$ 4.57) & \textbf{81.49($\pm$ 1.52)} & 70.77($\pm$ 1.00)  & 62.81($\pm$ 2.61) & 1.630($\pm$ 2.516)\\
\supervisedcontexttask & \textbf{91.54($\pm$ 3.52)} & \textbf{85.71($\pm$ 2.92)} & 81.23($\pm$ 1.94) & \textbf{71.36($\pm$ 1.05)} & 62.75($\pm$ 2.27) & 1.806($\pm$ 2.340)\\ 
%\hline
\Xhline{2\arrayrulewidth}
%\vspace{-10mm}
\end{tabular}
}
\end{table*}
%\vspace{-3mm}
\paragraph{GNN Backbone} The GNN architecture also plays a role in graph pretraining. In~\citet{hu2019strategies}, the results show that pretraining on GNN variants like GIN~\citep{xu2018powerful} would improve the performance on downstream tasks, while the performance with architectures like GAT~\citep{velivckovic2017graph} would actually get worse performance with pretraining. As the GNNs based on 1-Weisfeiler-Lehman (WL) test have similar representation power~\citep{xu2018powerful} bounded by the Weisfeiler-Lehman isomorphism check~\citep{shervashidze2011weisfeiler}, we consider the two representative GNN architectures, namely the \textbf{GIN}~\citep{xu2018powerful} and \textbf{GraphSage}~\citep{hamilton2017representation}. They have shown benefits with graph pretraining in ~\citet{hu2019strategies}.
%\vspace{-3mm}
\paragraph{Pretraining dataset} In natural language pretraining, researchers observed a significant performance boost due to self-supervised pretraining on large-scale data, that is, the larger the pretraining dataset is, the better the downstream performance it is ~\citep{raffel2019exploring}. %We hope to obtain similar outstanding performance improvement as a result of pretraining with large scale datasets for graph pretraining. 
Inspired by this success in natural language processing, we test the algorithms on two unlabeled pretraining datasets with different scales.

\begin{itemize}[noitemsep,topsep=1pt,parsep=1pt,partopsep=0pt, leftmargin=*]
	\item \textbf{ZINC15~\citep{sterling2015zinc}}: ZINC15 contains 2 million molecules. This dataset was preprocessed following ~\citet{hu2019strategies}.
	\item \textbf{SAVI ~\citep{patel2020synthetically}}: The SAVI dataset contains about 1 billion molecules, which are significantly larger than ZINC15. To the best of our knowledge, it has never been used for pretraining tasks before. This dataset contains drug-like molecules synthesized by computer simulated reactions. 
	\end{itemize}

Additionaly, we used \textbf{ChEMBL~\citep{gaulton2012chembl}} as the supervised datasets. Different from the above ZINC15 and SAVI dataset which are only used for self-supervised pretraining, this dataset contains 500k drug-able molecules with 1,310 prediction target labels from bio-activity assays for drug discovery. Thus like in~\citet{hu2019strategies} we only leverage it for \emph{supervised} pretraining. 
\paragraph{Data split on downstream tasks} The downstream tasks for molecular domain we used are 5 benchmark datasets from MoleculeNet~\citep{wu2018moleculenet} (See Appendix~\ref{sec:molecular_benchmarks} for more details). The train/valid/test sets are split with ratio 8:1:1. For molecule domain, the random split is not the most meaningful way to assess the performance, because the real-world scenarios often require generalization ability on out-of-distribution samples. So we consider the following ways to split the data:
\begin{itemize}[noitemsep,topsep=1pt,parsep=1pt,partopsep=0pt, leftmargin=*]
	\item \textbf{Scaffold Split~\citep{hu2019strategies, ramsundar2019deep}} This strategy first sorts the molecules according to the scaffold (e.g. molecule structure), and then partition the sorted list into train/valid/test splits consecutively. Therefore, the molecules in train and test sets are most different ones according to their molecule structure.  Note this strategy would yield deterministic data splits.
	\item \textbf{Balanced Scaffold Split~\citep{bemis1996properties, rong2020self}} This strategy introduces the randomness in the sorting and splitting stages above, thus one can run on splits with different random seeds and report the average performance to lower the evaluation variance.
\end{itemize}

We choose balanced scaffold as our major evaluation configuration, because it allows us to evaluate the algorithm on multiple data splits while maintaining the ability to evaluate out of distribution samples (e.g. assess generalization ability). Evaluating on one single split (such as scaffold split) can be subject to bias due to one specific split, leading to higher variance in evaluation. 
%\vspace{-3mm}
%x
\begin{table*}[t]
\small
%\vspace{-3mm}
\centering
\caption{Unsupervised + \textbf{Basic feature + Scaffold}. No pretrain has an average accuracy of 68.7\% over all benckmark datasets. \label{tab:Scaffold + basic feature + unsupervised}}
%\vspace{-3mm}
\vspace{2mm}
\resizebox{\textwidth}{!}{%
\begin{tabular}{lcccccr}
\Xhline{2\arrayrulewidth}
%\hline
Methods               & \textbf{BBBP}                & \textbf{BACE}             & \textbf{TOX21}            & \textbf{TOXCAST}              & \textbf{SIDER} &    \textbf{AVE GAIN}    \\ 
%\hline
\Xhline{2\arrayrulewidth}
\nopretrain              & 69.62($\pm$ 1.05)            & 75.77($\pm$4.29)          & 75.52($\pm$0.67)          & 63.67($\pm$0.32)              & 59.07($\pm$1.13)  &   0($\pm$1.492)\\
\masktask                & 68.70($\pm$2.16)             & 76.95($\pm$0.12)          & 75.88($\pm$0.60)          & 64.11($\pm$0.38)              & 61.29($\pm$0.87)  &   0.656($\pm$0.826)\\
\contexttask             & 69.41($\pm$1.44)             & \textbf{81.96}($\pm$0.72)          & 75.49($\pm$0.75)          & 63.48($\pm$0.31)              & \textbf{62.27($\pm$0.90)}  &   1.792($\pm$0.824)\\
%\supervisedtask          & 68.96($\pm$0.64)             & 76.30($\pm$1.30)          & 76.64($\pm$0.39)          & 66.07($\pm$0.22)              & 61.97($\pm$0.96) \\
%\supervisedmasktask      & \textbf{71.41($\pm$0.67)}    & 84.59($\pm$0.35)          & 79.13($\pm$0.29)          & 65.32($\pm$0.37)              & 62.12($\pm$0.19) \\
%\supervisedcontexttask   & 69.63($\pm$0.25)             & \textbf{83.34($\pm$0.67)} & 78.11($\pm$0.28)          & 66.15($\pm$0.48)              & 63.48($\pm$0.43) \\ 
%\hline
\Xhline{2\arrayrulewidth}
\end{tabular}
}
\end{table*}

\begin{table*}[t]
% \small
%\vspace{-5mm}
\centering
\caption{Supervised + \textbf{Basic feature + Scaffold}. No pretrain has an average accuracy of 68.7\% over all benckmark datasets. \label{tab:Scaffold + basic feature + supervised}}
\vspace{2mm}
\resizebox{\textwidth}{!}{%
\begin{tabular}{lcccccr}
%\hline
\Xhline{2\arrayrulewidth}
Methods                & \textbf{BBBP}                & \textbf{BACE}             & \textbf{TOX21}            & \textbf{TOXCAST}              & \textbf{SIDER}   &    \textbf{AVE GAIN}  \\ 
%\hline
\Xhline{2\arrayrulewidth}
\nopretrain              & 69.62($\pm$ 1.05)            & 75.77($\pm$4.29)          & 75.52($\pm$0.67)          & 63.67($\pm$0.32)              & 59.07($\pm$1.13)   &  0($\pm$1.492)\\
%\masktask                & 68.70($\pm$2.16)             & 76.95($\pm$0.12)          & 75.88($\pm$0.60)          & 64.11($\pm$0.38)              & 61.29($\pm$0.87) \\
%\contexttask             & 69.41($\pm$1.44)             & 81.96($\pm$0.72)          & 75.49($\pm$0.75)          & 63.48($\pm$0.31)              & 62.27($\pm$0.9)  \\
\supervisedtask          & 68.96($\pm$0.64)             & 76.30($\pm$1.30)          & 76.64($\pm$0.39)          & 66.07($\pm$0.22)              & 61.97($\pm$0.96)  &   1.258($\pm$0.702)\\
\supervisedmasktask      & \textbf{71.41($\pm$0.67)}    & 84.59($\pm$0.35)          & \textbf{79.13($\pm$0.29)} & 65.32($\pm$0.37)              & 62.12($\pm$0.19) &    3.784($\pm$0.374)\\
\supervisedcontexttask   & 69.63($\pm$0.25)             & \textbf{83.34($\pm$0.67)} & 78.11($\pm$0.28)          & \textbf{66.15($\pm$0.48)}     & \textbf{63.48($\pm$0.43)} &   3.412($\pm$0.422)\\ 
%\hline
\Xhline{2\arrayrulewidth}
%\vspace{-12mm}
\end{tabular}

}
\end{table*}
\subsection{Experiment protocol}
\label{sec:protocol}
As the total number of configurations for the entire pipeline can be combinatorially large which is not practical for us to exhaustively experiment with all of them, we design our protocol with a pairwise comparison principle. Specifically, we first anchor a \emph{vanilla configuration} with a certain design choice of combination for each stage. To study the effect of each stage on the pretraining effectiveness, we vary the design choice one stage at a time compared to the \emph{vanilla configuration}. %In total we would run $N$ sets of experiments where $N$ is the number of varying stages we considered. Note that in each set of experiments we will still have multiple experiments for each possible design choices we consider for that stage.

%By pairwise comparison between each set of variations and the \emph{vanilla configuration}, we can see how the design choices affect the outcome of the pretraining, and in which scenario will the pretraining help most for the downstream tasks. However, whenever feasible, we can also analyze the effect of joint variations on multiple stages by comparing the results of them together. We will present these results, our analysis and hypothesis in the next section. 

For all these experiments, to assess the effectiveness of graph pretraining, we report the \textbf{ROC-AUC} on downstream tasks as well as the relative \textbf{average gain} over all downstream datasets with and without pretraining. For each setting we will report the mean and standard deviation (in parenthesis) over three runs with different random seeds. We tune the model on downstream tasks with the validation set, and report the evaluation metric on the test set using the model with best validation performance. For each setup, we report the average performance obtained with three random seeds. We tune the learning rate in $\{1e^{-4}, 5e^{-4}, 1e^{-3}, 5e^{-3}, 1e^{-2}, 5e^{-2}, 1e^{-1}\}$ for each setup \emph{individually} and select the one with best validation performance. For GNNs we fix the hidden dimension to 300 and number of layers to 5.

%\vspace{-3mm}
\section{Results}
\label{sec:results}
%\vspace{-3mm}
In this section, we present the results and discussions for a set of experiments designed with the protocols in \Secref{sec:protocol}. \tabref{tab:summary-table} summarizes the experimental configurations for each following table. We will elaborate on them in the following sections. 
Due to space limit, we defer our investigation on different GNN architectures to appendix (\secref{sec：gnn}).

%\input{summary_table}
%\vspace{-2mm}
\subsection{Vanilla configuration}
%\vspace{-1mm}
We choose the \emph{vanilla configuration} with the settings from existing works~\citep{hu2019strategies, rong2020self}. Specifically, we use the \textbf{rich feature} with \textbf{GIN} backbone, pretrained on \textbf{ZINC15} when pretraining is applied, and evaluate on the \textbf{Balanced Scaffold Split} for downstream tasks. One important baseline is without pretraining. For the ease of comparison, we include the results without pretraining in each table.
%\vspace{-2mm}
\subsection{Self-supervised pretraining objectives}
We compare the results pretrained with different self-supervised pretraining objectives. As is presented in \Secref{sec:design_choice}, we consider four representative types of pretraining objectives. For the ease of comparing the performance, we only consider one objective at a time, instead of mixing different pretraining objectives to obtain a multi-task pretrained model. 
\tabref{tab:unsupervised-obj} shows the performance on downstream molecule property prediction benchmarks with models initialized from different pretraining objectives. The relative average gain compared to the one without pretraining is not statistically significant, i.e., not larger than the standard deviations of multiple runs. All the four different objectives obtain similar gains/loses regardless of very different designs. To fully understand the effect of self-supervised pretraining on molecule representation, we further investigate the performance of different pretraining objectives in combination with other factors, such as input features or data splits, as described in the following sections. 
%\vspace{-2mm}
\subsection{Supervised pretraining objectives}
In addition to the self-supervised objectives, we study the potential benefits with supervised pretraining. Unlike the self-supervised setting where the molecule graphs themselves are used for pretraining, the supervised pretraining requires extra cost of data labeling, and thus is not scalable for large scale pretraining. In this paper, we present the results with supervised pretraining alone, as well as the joint pretraining. e.g. pretrain with self-supervised objective and followed by supervised pretraining, in \tabref{tab:super-obj}. We can see with the supervised pretraining, one can improve the downstream performance, which aligns with the observation from \citet{hu2019strategies}. Our hypothesis is that, supervised pretraining is helpful when the pretraining tasks are closely aligned with the downstream tasks. In particular, the bio-activity labels provided by ChEMBL is highly related to the drug discovery purpose and drug discovery properties are the major topics evaluated in the downstream tasks. %\tabref{tab:super-obj} shows that the downstream tasks that are directly measuring biophysical property for drug discovery, such as toxicity, obtain the largest performance improvement from ChEMBL pretraining (e.g. TOXCAST and TOX21). %On the contrary, tasks that do not directly measure drug-able property such as BBBP (predict blood-brain barrier) achieve less gain.  
Therefore, the positive correlation between the pretraining supervision and downstream tasks contribute the most to the performance improvement of downstream tasks. %Based on this fact, the requirement for additional labels and the close alignment between pretraining and downstream tasks can limite the useage of supervised pretraining. 
%\vspace{-3mm}
\subsection{Data split on downstream tasks}
Molecular data is usually diverse and limited, so chemists are particularly interested in the generalization ability of GNNs on out of distribution data. Also due to the same reason (i.e. limited and diverse data), the variance in performance of different splits is significant, which poses challenges on robust evaluation. In \emph{vanilla configuration} we use the balanced scaffold split, and here we show additional results with the \textbf{scaffold} split, which is a deterministic data split that makes the train/valid/test set differ from each other the most.  \tabref{tab:split} and \tabref{tab:feature-scaffold-richfeature-supervised} respectively present the results using scaffold split with self-supervised without and with additional supervised pretraining. Compared with \tabref{tab:unsupervised-obj} and \tabref{tab:super-obj}, it is clear to see that \tabref{tab:split} and \tabref{tab:feature-scaffold-richfeature-supervised} have significantly lower ROC-AUC. Specifically the AUC drops 6.2\% on average for all benchmarks without pretraining. On the other hand, we can see if we compare \tabref{tab:split} with \tabref{tab:unsupervised-obj}, or \tabref{tab:feature-scaffold-richfeature-supervised} with \tabref{tab:super-obj} respectively, the gain of pretraining is more significant on the \textbf{scaffold} split. We 
speculate the reason for the improvement of scaffold split is that the initialization of neural network parameters (e.g. from pretraining) are typically critical for the out-of-distribution generalization (e.g. scaffold split). Similar observations have also been studied in the meta-learning literature~\citep{finn2017model}.  Although the gain with supervised pretraining is significant in \tabref{tab:feature-scaffold-richfeature-supervised},  the effect of self-supervised pretraining is mixed in \tabref{tab:split}. This indicates the effectiveness of self-supervised pretraining on scaffold split is not significant enough to claim ``very helpful''.  
%\vspace{-3mm}
\subsection{Graph features}
%\vspace{-2mm}
So far we have presented the results with rich features. Now we want to see how those basic features used in~\citet{hu2019strategies} affect the outcome. \tabref{tab:Balanced scaffold + basic feature + unsupervised} and \tabref{tab:Balanced scaffold + basic feature + supervised} show the test ROC-AUC (\%) performance with {\it basic features} on the balanced scaffold splits using self-supervised or supervised pretraining objectives, respectively. \tabref{tab:Scaffold + basic feature + unsupervised} and \tabref{tab:Scaffold + basic feature +  supervised} show the same results but on scaffold split. %instead. 

In a nutshell, without pretraining, rich features lead to an average gain of $1.3\%$ and $3.1\%$ over basic features using balanced scaffold split and scaffold split, respectively. Specifically, it achieves $76.7\%$ vs $78.0\%$ for balanced scaffold split, and $68.7\%$ vs $71.8\%$ on scaffold split. The gain brought by the rich features are more significant than the ones with different self-supervised pretraining objectives. \tabref{tab:Balanced scaffold + basic feature + unsupervised} to \tabref{tab:Scaffold + basic feature +  supervised} show that pretraining has more positive impact when basic features are used. In particular, the self-pretraining with context prediction shows significant gains especially in the {scaffold split} setting. However, the gain diminishes when careful feature engineering are applied to the downstream tasks (use rich feature in \emph{vanilla configuration}). The supervised pretraining continues the significant gain under these settings, which shows the consistency and reliability of the situation with the labeled and downstream-task-aligned supervisions. %Specifically, the gain brought by supervised pretraining can be more significant than the rich feature engineering alone. 
%\vspace{-7.5mm}
\subsection{Pretraining datasets}
\label{sec:exp_data_size}
%\vspace{-2mm}
\begin{table*}[t]
\centering
%\vspace{-5mm}
\caption{\textbf{Large scale pretraining data} with balanced scaffold split. No pretraining gets an average AUC of 78.0\%. \label{tab:savi-balanced-split}}
\vspace{2mm}
\resizebox{\textwidth}{!}{%
\begin{tabular}{llllllr}
%\hline
\Xhline{2\arrayrulewidth}
Methods        & \textbf{BBBP}  & \textbf{BACE}  & \textbf{TOX21} & \textbf{TOXCAST} & \textbf{SIDER}   & \textbf{AVE GAIN}\\ 
%\hline
\Xhline{2\arrayrulewidth}
\nopretrain      & 92.23($\pm$3.07) & 87.43($\pm$1.63)     & 79.2($\pm$1.99)  & 69.13($\pm$0.55) & 61.92($\pm$0.89) & 0($\pm$1.626)\\
\masktask     & 92.33($\pm$2.08) & 87.22($\pm$1.79) & 79.12($\pm$1.62) & 69.47($\pm$0.65)   & 61.24($\pm$1.94)  &  -0.106($\pm$1.616)  \\
\contexttask  &  93.32($\pm$0.53) & 87.77($\pm$2.94) & 79.18($\pm$2.48) & 70.13($\pm$0.56) & 62.24($\pm$2.65)  &  0.546($\pm$1.832) \\ 
\Xhline{2\arrayrulewidth}
\rowcolor{Gray}
\supervisedmasktask     & 93.23($\pm$3.02) & 86.39($\pm$1.67) & 81.89($\pm$1.58) & 71.77($\pm$0.50)  & 63.73($\pm$2.20)   &  1.420($\pm$1.794) \\
\rowcolor{Gray}
\supervisedcontexttask & 92.55($\pm$2.93) & 87.76($\pm$1.87) & 82.19($\pm$1.58) & 72.91($\pm$0.71) & 62.44($\pm$0.45)  &  1.588($\pm$1.508) \\
%\hline
\Xhline{2\arrayrulewidth}
 %\vspace{-10mm}
\end{tabular}%
}
\end{table*}

\begin{table*}[t]
% \small
\caption{\textbf{Large scale pretraining} data with scaffold split. No pretraining gets an average AUC of 71.8\%. \label{tab:SAVI-scaffold-split}}
%\vspace{-3mm}
\vspace{2mm}
\centering
\resizebox{\textwidth}{!}{%
\begin{tabular}{llllllr}
%\hline
\Xhline{2\arrayrulewidth}
Methods               & \textbf{BBBP}  & \textbf{BACE} & \textbf{TOX21} & \textbf{TOXCAST} & \textbf{SIDER}  &  \textbf{AVE GAIN}\\ 
%\hline
\Xhline{2\arrayrulewidth}
\nopretrain             & 74.83($\pm$0.73) & 80.10($\pm$0.42) & 75.86($\pm$0.58)  & 65.95($\pm$0.15) & 62.30($\pm$1.14)  & 0($\pm$0.604)\\
\masktask               & 73.81($\pm$1.82) & 81.90($\pm$1.59) & 74.94($\pm$0.05) & 66.95($\pm$0.12)  & 62.93($\pm$0.34)  &  0.298($\pm$0.784) \\
\contexttask            & 74.32($\pm$0.85) & 83.93($\pm$0.24) & 74.42($\pm$0.19) & 67.01($\pm$0.29)  & 64.83($\pm$0.45)  & 1.094($\pm$0.404)\\ 
\Xhline{2\arrayrulewidth}
\rowcolor{Gray}
\supervisedmasktask     & 73.32($\pm$0.60) & 83.38($\pm$1.05) & 78.59($\pm$0.09) & 67.01($\pm$0.18)  & 65.40($\pm$0.12)  & 1.732($\pm$0.408)\\
\rowcolor{Gray}
\supervisedcontexttask  & 74.38($\pm$0.93) & 86.33($\pm$0.16) & 78.16($\pm$0.25) & 68.71($\pm$0.07)  & 62.22($\pm$0.48)  & 2.152($\pm$0.378)\\
%\hline
\Xhline{2\arrayrulewidth}
%\vspace{-10mm}
\end{tabular}%
}
\end{table*}

As observed in natural language processing domain, more text pretraining data lead to better downstream performance. Intuitively this can be true for molecule representation domain as well, so we run a new set of experiments with the model pretrained on SAVI dataset, which is about 500 times larger than the ZINC15 dataset we used in the above result sections. We present the results pretrained on SAVI dataset using balanced scaffold split or scaffold split in \tabref{tab:savi-balanced-split} and \tabref{tab:SAVI-scaffold-split}, respectively. Other configurations are the same as the \emph{vanilla configuration}. 

Compared with the performance on ZINC15, the SAVI pretraining data does not lead to a significant improvement either on balanced scaffold split  (\tabref{tab:unsupervised-obj} vs \tabref{tab:savi-balanced-split}) or scaffold split (\tabref{tab:split} or \tabref{tab:SAVI-scaffold-split}). Similarly, the self-supervised pretraining objectives lead to negligible gain on downstream task performance, while the supervised one still achieves a clear gain. 

\begin{wrapfigure}{r}{0.59\textwidth}
 \vspace{-9mm}
  \begin{center}
    \includegraphics[width=0.36\textwidth]{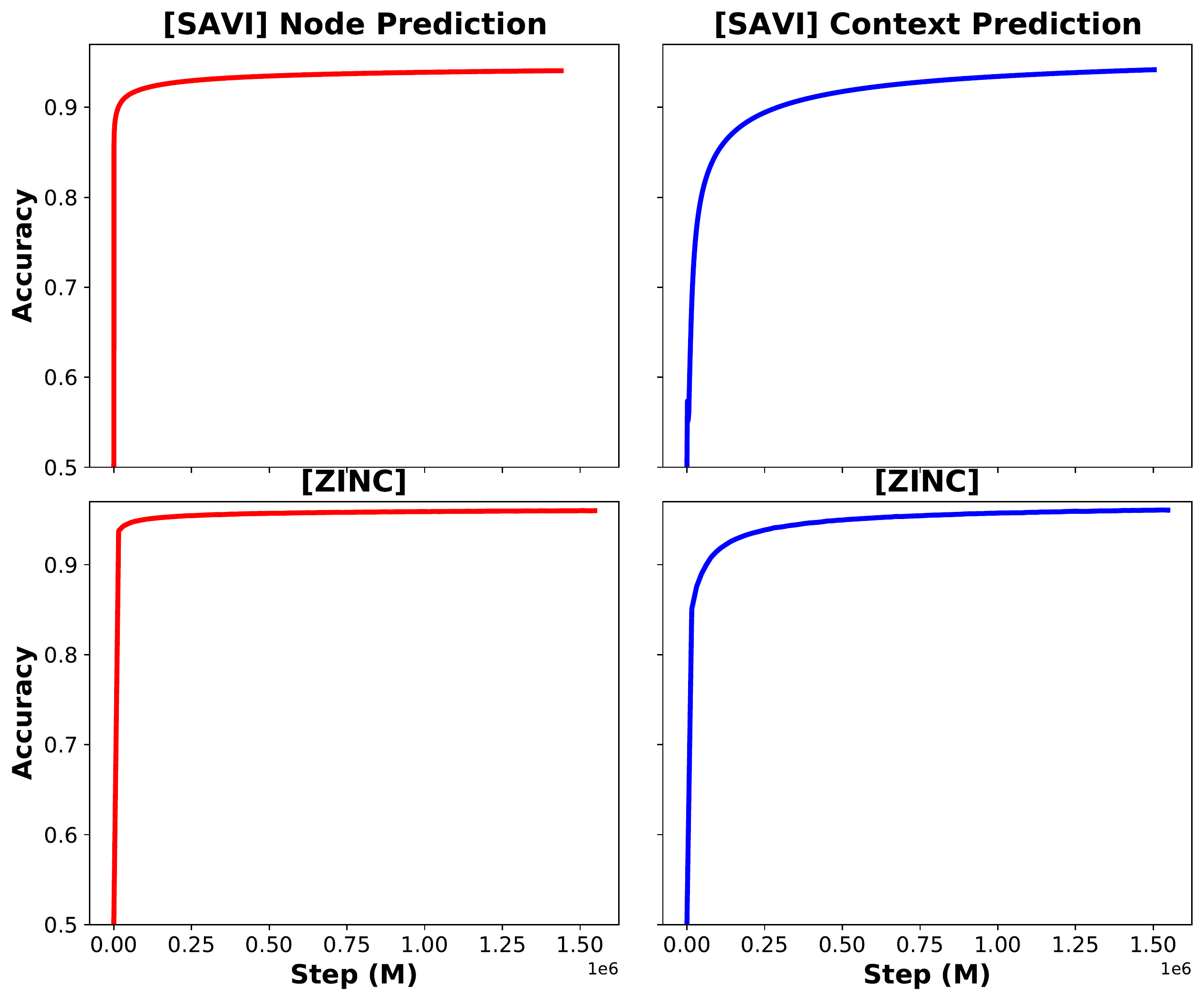}
  \end{center}
  \vspace{-3mm}
 \caption{Pretraining accuracy on ZINC15 or SAVI datasets with node prediction or context prediction objectives. \label{fig:pretraining}}
 \vspace{-3mm}
\end{wrapfigure}

As the result is counterintuitive, we further investigate the reason behind it by inspecting the pretraining performances with different training objectives on both ZINC15 and SAVI datasets. We plot the curve of accuracy growth with the number of training steps iterated. We can see from \Figref{fig:pretraining} that in all settings the pretraining accuracy grows above 90\% quickly after only 0.1 to 0.2M steps and also converges quickly. Given that the model gets very high accuracy without even going through 1 epoch of the SAVI dataset, it is expected that the larger training data like SAVI may not provide more learning signals for the model, and partially explains why more molecules wouldn't help significantly in this case. Furthermore, these figures might suggest several reasons of why the self-supervised pretraining may not be very effective in some situations:
\begin{itemize}[noitemsep,topsep=2pt,parsep=6pt,partopsep=2pt, leftmargin=*]
	\item \textbf{Tasks are easy.} Some of the self-pretraining tasks for molecules might be easy, so that model learns less useful information from pretraining. For example in the masked node prediction case, the model is expected to predict the atomic number from a vocabulary with less than 100 candidate atoms. Furthermore, due to the valence constraints, the graph topology may already exclude most of the wrong atoms. As a comparison, the vocabulary size for text pretraining may be 100k or even higher. Some structured prediction tasks like context prediction might be hard, but due to the difficulty of structured prediction itself and the proposal for high quality negative examples for contrastive learning, it can still be challenging for downstream task improvements. Other strategies like motif prediction can be achieved by subgraph matching, which can be easy for GNN that intrinsically does the graph isomorphism test.
	\item \textbf{Data lacks diversity.} Due to biophysical and functional requirements, molecules share many common sub-structures, e.g., functional motifs. Hence, molecules may not be as diversified as text data. This is why the model learns to generalize quickly within the training distribution. 
	\item \textbf{2D Structure is not enough to infer functionality.} Some important biophysical properties (such as 3D structure, chirality) are barely reflected in the 2D-feature-based pretraining (e.g., using smiles or 2D graph features). For example, the molecules with the same chemical formula and 2D feature, can have very different chirality, which leads to quite different toxicity \citep{smith2009chiral} (e.g. flipped toxicity labels). This is not captured in the current GNN pretraining frameworks that we considered.
\end{itemize}
%\vspace{-1mm}

%\begin{wrapfigure}{r}{0.5\textwidth}
%\begin{figure}%[t]
%\centering
%\includegraphics[width=0.46\textwidth]{figs/pretrain-4.pdf}
%\caption{Pretraining accuracy on ZINC15 or SAVI datasets with node prediction or context prediction objectives. \label{fig:pretraining}}
%\vspace{-1mm}
%\end{figure}
%\end{wrapfigure}

%\subsection{GNN architecture} 
%We also investigate GNN
% \vspace{-4mm}
\subsection{Hyper-parameters for downstream tasks}
We also find that hyper-parameters for downstream tasks are critical for the their performance that their choices may change the conclusion of the effectiveness of pretraining in some settings.
% For example, learning rate is important. 
We can take the learning rate as an example.
As the models initialized from scratch and pretraining may have different scales, the most suitable learning rate required for downstream tasks may also be different. Without tuning learning rate extensively, we may reach a misleading conclusion.
In particular, when we adopt the default learning rate for reproducing the existing success of pretraining in \tabref{tab:feature-balancedscaffold-basicfeature-wrong} of Appendix~\ref{sec:reproduce}, we indeed observe the advantage of pretraining. However, if we follow our procedures (e.g. extensive search learning rate and averaging over three splits), the resulting \tabref{tab:unsupervised-obj} and \tabref{tab:super-obj} indicate no performance gain by pretraining. So we suggest that the evaluation of pretraining should consider the hyper-parameter tuning and averaging over different splits.

%\vspace{-1mm}
\section{Summary and takeaways}
\label{sec:summary}
%\vspace{-1mm}
Based on our experiments in \secref{sec:results}, we present our takeaways by empirically summarizing our conjectures on when the pretraining would/would not help the molecular representation learning.
%\vspace{-3mm}
\paragraph{When pretraining might help} We find it typically helps 1) if we can have the supervised pretraining with target labels that are aligned with the downstream tasks. However, getting large amount of high-quality and relevant supervision is not always feasible; 2) if the high quality hand-crafted features are absent. However, it seems that the gain obtained by self-supervised pretraining is not as significant as these high quality hand-crafted features based on our current studies; 3) if the downstream train, valid and test dataset distributions are substantially different.
%\vspace{-3mm}
\paragraph{When the gain diminishes?} In some situations the gain of pretraining might diminish 1) if we already have the high quality hand-crafted features (e.g. rich features described in \secref{sec:design_choice}); 2) if we don't have the highly relevant supervisions. As shown in \secref{sec:exp_data_size}, many self-supervised pretraining tasks might be too easy for the model to learn meaningful embedding; 3) if the downstream data splits are balanced; 4) if the self-supervised learning dataset lacks diversity, despite its scale.
\paragraph{Why pretraining may not help in some cases?} In our paper we pretrained a GNN on a much larger dataset (SAVI) than before, hoping to replicate gain of pretraining like in NLP domain. However, we do not obtain the expected gain. The pretraining accuracy curve (\Figref{fig:pretraining}) provides some potential explanations of why pretraining may not work: some of the pretraining accuracy curve grows above $95\%$+ quickly and converges fast, unlike pretraining in NLP which keeps growing to $70\%$ and hardly plateaus. This suggests that some of the pretraining methods like masked node label prediction might be easy (as the vocabulary size is much smaller compared to NLP) and therefore transfer less knowledge for downstream tasks.

%\vspace{-3mm}
\section{Limitations of current study}
\label{sec:limitation}
%\vspace{-3mm}
Although we have tried our best to design a comprehensive study on the effectiveness of graph pretraining for molecular representation, there are still limitations we want to point out. Due to the limited time and resources we have, we are not able to fully cover the whole picture of the current pretraining paradigm in graph neural networks. Nevertheless, we list them here in hope of preventing the over-generalization of our conclusion.
\begin{itemize}[leftmargin=*,nosep]
    \item \textit{Distribution of graphs}. Our study focuses on pretraining for small molecule graph inductive representation learning. Recently there are works on pretraining transductive representation learning~\citep{you2020does} on large graphs~\citep{hu2020gpt}, where our conclusion may not be directly extended to these cases. 
    \item \textit{Graph architectures}. GNN is a popular research field where many new architectures with probable expressiveness are/will be proposed. The results we have shown are on two representative 1-WL GNNs. It can be possible that the latest advances of deep GNN~\citep{liu2020towards} and Transformer-based GNN~\citep{ying2021transformers, chen2022structure, kim2022pure} might yield different results.
    \item \textit{Learning objectives}. Although we have presented results with different types of self-supervised losses, there are still many variants of each type that we did not explore, like different variants~\citep{sun2021mocl, xia2022simgrace} of contrastive learning. Also, multi-task learning of different self-supervised objectives might be another direction for further exploration.
    \item \textit{Downstream datasets}. We obtained our conclusion mainly on the datasets from MoleculeNet~\citep{wu2018moleculenet}. Datasets like Alchemy~\citep{chen2019alchemy} and drug-target Interaction~\citep{liu2021pre} may show different results.
\end{itemize}

 \clearpage

\nocite{langley00}

%\bibliography{example_paper}
%\bibliographystyle{icml2022}
%\bibliographystyle{unsrt}
%\bibliographystyle{plainnat}
%\bibliographystyle{unsrtnat}
\bibliographystyle{plainnat}
\bibliography{example_paper.bib}

\begin{thebibliography}{37}
\providecommand{\natexlab}[1]{#1}
\providecommand{\url}[1]{\texttt{#1}}
\expandafter\ifx\csname urlstyle\endcsname\relax
  \providecommand{\doi}[1]{doi: #1}\else
  \providecommand{\doi}{doi: \begingroup \urlstyle{rm}\Url}\fi

\bibitem[Bemis and Murcko(1996)]{bemis1996properties}
Guy~W Bemis and Mark~A Murcko.
\newblock The properties of known drugs. 1. molecular frameworks.
\newblock \emph{Journal of medicinal chemistry}, 39\penalty0 (15):\penalty0
  2887--2893, 1996.

\bibitem[Brown et~al.(2020)Brown, Mann, Ryder, Subbiah, Kaplan, Dhariwal,
  Neelakantan, Shyam, Sastry, Askell, et~al.]{brown2020language}
Tom~B Brown, Benjamin Mann, Nick Ryder, Melanie Subbiah, Jared Kaplan, Prafulla
  Dhariwal, Arvind Neelakantan, Pranav Shyam, Girish Sastry, Amanda Askell,
  et~al.
\newblock Language models are few-shot learners.
\newblock \emph{arXiv preprint arXiv:2005.14165}, 2020.

\bibitem[Chen et~al.(2022)Chen, O’Bray, and Borgwardt]{chen2022structure}
Dexiong Chen, Leslie O’Bray, and Karsten Borgwardt.
\newblock Structure-aware transformer for graph representation learning.
\newblock In \emph{International Conference on Machine Learning}, pages
  3469--3489. PMLR, 2022.

\bibitem[Chen et~al.(2019)Chen, Chen, Hsieh, Lee, Liao, Liao, Liu, Qiu, Sun,
  Tang, et~al.]{chen2019alchemy}
Guangyong Chen, Pengfei Chen, Chang-Yu Hsieh, Chee-Kong Lee, Benben Liao,
  Renjie Liao, Weiwen Liu, Jiezhong Qiu, Qiming Sun, Jie Tang, et~al.
\newblock Alchemy: A quantum chemistry dataset for benchmarking ai models.
\newblock \emph{arXiv preprint arXiv:1906.09427}, 2019.

\bibitem[Chen et~al.(2020)Chen, Kornblith, Norouzi, and Hinton]{chen2020simple}
Ting Chen, Simon Kornblith, Mohammad Norouzi, and Geoffrey Hinton.
\newblock A simple framework for contrastive learning of visual
  representations.
\newblock In \emph{International conference on machine learning}, pages
  1597--1607. PMLR, 2020.

\bibitem[Devlin et~al.(2018)Devlin, Chang, Lee, and Toutanova]{devlin2018bert}
Jacob Devlin, Ming-Wei Chang, Kenton Lee, and Kristina Toutanova.
\newblock Bert: Pre-training of deep bidirectional transformers for language
  understanding.
\newblock \emph{arXiv preprint arXiv:1810.04805}, 2018.

\bibitem[Finn et~al.(2017)Finn, Abbeel, and Levine]{finn2017model}
Chelsea Finn, Pieter Abbeel, and Sergey Levine.
\newblock Model-agnostic meta-learning for fast adaptation of deep networks.
\newblock In \emph{International Conference on Machine Learning}, pages
  1126--1135. PMLR, 2017.

\bibitem[Gaulton et~al.(2012)Gaulton, Bellis, Bento, Chambers, Davies, Hersey,
  Light, McGlinchey, Michalovich, Al-Lazikani, et~al.]{gaulton2012chembl}
Anna Gaulton, Louisa~J Bellis, A~Patricia Bento, Jon Chambers, Mark Davies,
  Anne Hersey, Yvonne Light, Shaun McGlinchey, David Michalovich, Bissan
  Al-Lazikani, et~al.
\newblock Chembl: a large-scale bioactivity database for drug discovery.
\newblock \emph{Nucleic acids research}, 40\penalty0 (D1):\penalty0
  D1100--D1107, 2012.

\bibitem[Hamilton et~al.(2017)Hamilton, Ying, and
  Leskovec]{hamilton2017representation}
William~L Hamilton, Rex Ying, and Jure Leskovec.
\newblock Representation learning on graphs: Methods and applications.
\newblock \emph{arXiv preprint arXiv:1709.05584}, 2017.

\bibitem[Hassani and Khasahmadi(2020)]{hassani2020contrastive}
Kaveh Hassani and Amir~Hosein Khasahmadi.
\newblock Contrastive multi-view representation learning on graphs.
\newblock In \emph{International Conference on Machine Learning}, pages
  4116--4126. PMLR, 2020.

\bibitem[He et~al.(2020)He, Fan, Wu, Xie, and Girshick]{he2020moco}
Kaiming He, Haoqi Fan, Yuxin Wu, Saining Xie, and Ross~B. Girshick.
\newblock Momentum contrast for unsupervised visual representation learning.
\newblock In \emph{CVPR}, pages 9726--9735, 2020.

\bibitem[Hu et~al.(2019)Hu, Liu, Gomes, Zitnik, Liang, Pande, and
  Leskovec]{hu2019strategies}
Weihua Hu, Bowen Liu, Joseph Gomes, Marinka Zitnik, Percy Liang, Vijay Pande,
  and Jure Leskovec.
\newblock Strategies for pre-training graph neural networks.
\newblock \emph{arXiv preprint arXiv:1905.12265}, 2019.

\bibitem[Hu et~al.(2020)Hu, Dong, Wang, Chang, and Sun]{hu2020gpt}
Ziniu Hu, Yuxiao Dong, Kuansan Wang, Kai-Wei Chang, and Yizhou Sun.
\newblock Gpt-gnn: Generative pre-training of graph neural networks.
\newblock In \emph{Proceedings of the 26th ACM SIGKDD International Conference
  on Knowledge Discovery \& Data Mining}, pages 1857--1867, 2020.

\bibitem[Kim et~al.(2022)Kim, Nguyen, Min, Cho, Lee, Lee, and
  Hong]{kim2022pure}
Jinwoo Kim, Tien~Dat Nguyen, Seonwoo Min, Sungjun Cho, Moontae Lee, Honglak
  Lee, and Seunghoon Hong.
\newblock Pure transformers are powerful graph learners.
\newblock \emph{arXiv preprint arXiv:2207.02505}, 2022.

\bibitem[Landrum(2016)]{landrum2016rdkit}
G~Landrum.
\newblock Rdkit: Open-source cheminformatics software, 2016.

\bibitem[Langley(2000)]{langley00}
P.~Langley.
\newblock Crafting papers on machine learning.
\newblock In Pat Langley, editor, \emph{Proceedings of the 17th International
  Conference on Machine Learning (ICML 2000)}, pages 1207--1216, Stanford, CA,
  2000. Morgan Kaufmann.

\bibitem[Liu et~al.(2020)Liu, Gao, and Ji]{liu2020towards}
Meng Liu, Hongyang Gao, and Shuiwang Ji.
\newblock Towards deeper graph neural networks.
\newblock In \emph{Proceedings of the 26th ACM SIGKDD International Conference
  on Knowledge Discovery \& Data Mining}, pages 338--348, 2020.

\bibitem[Liu et~al.(2021)Liu, Wang, Liu, Lasenby, Guo, and Tang]{liu2021pre}
Shengchao Liu, Hanchen Wang, Weiyang Liu, Joan Lasenby, Hongyu Guo, and Jian
  Tang.
\newblock Pre-training molecular graph representation with 3d geometry.
\newblock \emph{arXiv preprint arXiv:2110.07728}, 2021.

\bibitem[Patel et~al.(2020)Patel, Ihlenfeldt, Judson, Moroz, Pevzner, Peach,
  Tarasova, and Nicklaus]{patel2020synthetically}
Hitesh Patel, Wolf Ihlenfeldt, Philip Judson, Yurii~S Moroz, Yuri Pevzner,
  Megan Peach, Nadya Tarasova, and Marc Nicklaus.
\newblock Synthetically accessible virtual inventory (savi).
\newblock 2020.

\bibitem[Raffel et~al.(2019)Raffel, Shazeer, Roberts, Lee, Narang, Matena,
  Zhou, Li, and Liu]{raffel2019exploring}
Colin Raffel, Noam Shazeer, Adam Roberts, Katherine Lee, Sharan Narang, Michael
  Matena, Yanqi Zhou, Wei Li, and Peter~J Liu.
\newblock Exploring the limits of transfer learning with a unified text-to-text
  transformer.
\newblock \emph{arXiv preprint arXiv:1910.10683}, 2019.

\bibitem[Ramsundar et~al.(2019)Ramsundar, Eastman, Walters, and
  Pande]{ramsundar2019deep}
Bharath Ramsundar, Peter Eastman, Patrick Walters, and Vijay Pande.
\newblock \emph{Deep learning for the life sciences: applying deep learning to
  genomics, microscopy, drug discovery, and more}.
\newblock " O'Reilly Media, Inc.", 2019.

\bibitem[Rong et~al.(2020)Rong, Bian, Xu, Xie, Wei, Huang, and
  Huang]{rong2020self}
Yu~Rong, Yatao Bian, Tingyang Xu, Weiyang Xie, Ying Wei, Wenbing Huang, and
  Junzhou Huang.
\newblock Self-supervised graph transformer on large-scale molecular data.
\newblock \emph{arXiv preprint arXiv:2007.02835}, 2020.

\bibitem[Shervashidze et~al.(2011)Shervashidze, Schweitzer, Van~Leeuwen,
  Mehlhorn, and Borgwardt]{shervashidze2011weisfeiler}
Nino Shervashidze, Pascal Schweitzer, Erik~Jan Van~Leeuwen, Kurt Mehlhorn, and
  Karsten~M Borgwardt.
\newblock Weisfeiler-lehman graph kernels.
\newblock \emph{Journal of Machine Learning Research}, 12\penalty0 (9), 2011.

\bibitem[Smith(2009)]{smith2009chiral}
Silas~W Smith.
\newblock Chiral toxicology: it's the same thing… only different.
\newblock \emph{Toxicological sciences}, 110\penalty0 (1):\penalty0 4--30,
  2009.

\bibitem[Sterling and Irwin(2015)]{sterling2015zinc}
Teague Sterling and John~J Irwin.
\newblock Zinc 15--ligand discovery for everyone.
\newblock \emph{Journal of chemical information and modeling}, 55\penalty0
  (11):\penalty0 2324--2337, 2015.

\bibitem[Sun et~al.(2021)Sun, Xing, Wang, Chen, and Zhou]{sun2021mocl}
Mengying Sun, Jing Xing, Huijun Wang, Bin Chen, and Jiayu Zhou.
\newblock Mocl: data-driven molecular fingerprint via knowledge-aware
  contrastive learning from molecular graph.
\newblock In \emph{Proceedings of the 27th ACM SIGKDD Conference on Knowledge
  Discovery \& Data Mining}, pages 3585--3594, 2021.

\bibitem[Veli{\v{c}}kovi{\'c} et~al.(2017)Veli{\v{c}}kovi{\'c}, Cucurull,
  Casanova, Romero, Lio, and Bengio]{velivckovic2017graph}
Petar Veli{\v{c}}kovi{\'c}, Guillem Cucurull, Arantxa Casanova, Adriana Romero,
  Pietro Lio, and Yoshua Bengio.
\newblock Graph attention networks.
\newblock \emph{arXiv preprint arXiv:1710.10903}, 2017.

\bibitem[Wei et~al.(2021)Wei, Bosma, Zhao, Guu, Yu, Lester, Du, Dai, and
  Le]{wei2021finetuned}
Jason Wei, Maarten Bosma, Vincent~Y Zhao, Kelvin Guu, Adams~Wei Yu, Brian
  Lester, Nan Du, Andrew~M Dai, and Quoc~V Le.
\newblock Finetuned language models are zero-shot learners.
\newblock \emph{arXiv preprint arXiv:2109.01652}, 2021.

\bibitem[Wu et~al.(2021)Wu, Lin, Gao, Tan, Li, et~al.]{wu2021self}
Lirong Wu, Haitao Lin, Zhangyang Gao, Cheng Tan, Stan Li, et~al.
\newblock Self-supervised on graphs: Contrastive, generative, or predictive.
\newblock \emph{arXiv preprint arXiv:2105.07342}, 2021.

\bibitem[Wu et~al.(2018)Wu, Ramsundar, Feinberg, Gomes, Geniesse, Pappu,
  Leswing, and Pande]{wu2018moleculenet}
Zhenqin Wu, Bharath Ramsundar, Evan~N Feinberg, Joseph Gomes, Caleb Geniesse,
  Aneesh~S Pappu, Karl Leswing, and Vijay Pande.
\newblock Moleculenet: a benchmark for molecular machine learning.
\newblock \emph{Chemical science}, 9\penalty0 (2):\penalty0 513--530, 2018.

\bibitem[Xia et~al.(2022)Xia, Wu, Chen, Hu, and Li]{xia2022simgrace}
Jun Xia, Lirong Wu, Jintao Chen, Bozhen Hu, and Stan~Z Li.
\newblock Simgrace: A simple framework for graph contrastive learning without
  data augmentation.
\newblock In \emph{Proceedings of the ACM Web Conference 2022}, pages
  1070--1079, 2022.

\bibitem[Xie et~al.(2021)Xie, Xu, Zhang, Wang, and Ji]{xie2021self}
Yaochen Xie, Zhao Xu, Jingtun Zhang, Zhengyang Wang, and Shuiwang Ji.
\newblock Self-supervised learning of graph neural networks: A unified review.
\newblock \emph{arXiv preprint arXiv:2102.10757}, 2021.

\bibitem[Xu et~al.(2018)Xu, Hu, Leskovec, and Jegelka]{xu2018powerful}
Keyulu Xu, Weihua Hu, Jure Leskovec, and Stefanie Jegelka.
\newblock How powerful are graph neural networks?
\newblock \emph{arXiv preprint arXiv:1810.00826}, 2018.

\bibitem[Ying et~al.(2021)Ying, Cai, Luo, Zheng, Ke, He, Shen, and
  Liu]{ying2021transformers}
Chengxuan Ying, Tianle Cai, Shengjie Luo, Shuxin Zheng, Guolin Ke, Di~He,
  Yanming Shen, and Tie-Yan Liu.
\newblock Do transformers really perform bad for graph representation?
\newblock \emph{arXiv preprint arXiv:2106.05234}, 2021.

\bibitem[You et~al.(2020{\natexlab{a}})You, Chen, Sui, Chen, Wang, and
  Shen]{you2020graph}
Yuning You, Tianlong Chen, Yongduo Sui, Ting Chen, Zhangyang Wang, and Yang
  Shen.
\newblock Graph contrastive learning with augmentations.
\newblock \emph{Advances in Neural Information Processing Systems},
  33:\penalty0 5812--5823, 2020{\natexlab{a}}.

\bibitem[You et~al.(2020{\natexlab{b}})You, Chen, Wang, and Shen]{you2020does}
Yuning You, Tianlong Chen, Zhangyang Wang, and Yang Shen.
\newblock When does self-supervision help graph convolutional networks?
\newblock In \emph{International Conference on Machine Learning}, pages
  10871--10880. PMLR, 2020{\natexlab{b}}.

\bibitem[Zhu et~al.(2021)Zhu, Xu, Yu, Liu, Wu, and Wang]{zhu2021graph}
Yanqiao Zhu, Yichen Xu, Feng Yu, Qiang Liu, Shu Wu, and Liang Wang.
\newblock Graph contrastive learning with adaptive augmentation.
\newblock In \emph{Proceedings of the Web Conference 2021}, pages 2069--2080,
  2021.

\end{thebibliography}
\clearpage
\section*{Checklist}

% % %%% BEGIN INSTRUCTIONS %%%
%  The checklist follows the references.  Please
%  read the checklist guidelines carefully for information on how to answer these
%  questions.  For each question, change the default \answerTODO{} to \answerYes{},
%  \answerNo{}, or \answerNA{}.  You are strongly encouraged to include a {\bf
%  justification to your answer}, either by referencing the appropriate section of
%  your paper or providing a brief inline description.  For example:
%  \begin{itemize}
%  \item Did you include the license to the code and datasets? \answerYes{See Section~\ref{gen_inst}.}
%   \item Did you include the license to the code and datasets? \answerNo{The code and the data are proprietary.}
%   \item Did you include the license to the code and datasets? \answerNA{}
%  \end{itemize}
% % Please do not modify the questions and only use the provided macros for your
% answers.  Note that the Checklist section does not count towards the page
% limit.  In your paper, please delete this instructions block and only keep the
% Checklist section heading above along with the questions/answers below.
% %%% END INSTRUCTIONS %%%

\begin{enumerate}

\item For all authors...
\begin{enumerate}
  \item Do the main claims made in the abstract and introduction accurately reflect the paper's contributions and scope?
    \answerYes{}
  \item Did you describe the limitations of your work?
    \answerYes{}
  \item Did you discuss any potential negative societal impacts of your work?
    \answerYes{}. Though not really apply. 
  \item Have you read the ethics review guidelines and ensured that your paper conforms to them?
    \answerYes{}
\end{enumerate}

\item If you are including theoretical results...
\begin{enumerate}
  \item Did you state the full set of assumptions of all theoretical results?
    \answerNA{}
        \item Did you include complete proofs of all theoretical results?
    \answerNA{}
\end{enumerate}

\item If you ran experiments...
\begin{enumerate}
  \item Did you include the code, data, and instructions needed to reproduce the main experimental results (either in the supplemental material or as a URL)?
    \answerNo{}. 
    {\color{blue}We will prepare code soon.} 
  \item Did you specify all the training details (e.g., data splits, hyperparameters, how they were chosen)?
    \answerYes{}
        \item Did you report error bars (e.g., with respect to the random seed after running experiments multiple times)?
    \answerYes{}
        \item Did you include the total amount of compute and the type of resources used (e.g., type of GPUs, internal cluster, or cloud provider)?
    \answerYes{}
\end{enumerate}

\item If you are using existing assets (e.g., code, data, models) or curating/releasing new assets...
\begin{enumerate}
  \item If your work uses existing assets, did you cite the creators?
    \answerYes{}
  \item Did you mention the license of the assets?
    \answerNA{}
  \item Did you include any new assets either in the supplemental material or as a URL?
    \answerNA{}
  \item Did you discuss whether and how consent was obtained from people whose data you're using/curating?
    \answerYes{}
  \item Did you discuss whether the data you are using/curating contains personally identifiable information or offensive content?
    \answerNA{}
\end{enumerate}

\item If you used crowdsourcing or conducted research with human subjects...
\begin{enumerate}
  \item Did you include the full text of instructions given to participants and screenshots, if applicable?
    \answerNA{}
  \item Did you describe any potential participant risks, with links to Institutional Review Board (IRB) approvals, if applicable?
    \answerNA{}
  \item Did you include the estimated hourly wage paid to participants and the total amount spent on participant compensation?
   \answerNA{}
\end{enumerate}

\end{enumerate}

%%%%%%%%%%%%%%%%%%%%%%%%%%%%%%%%%%%%%%%%%%%%%%%%%%%%%%%%%%%%%%%%%%%%%%%%%%%%%%%
%%%%%%%%%%%%%%%%%%%%%%%%%%%%%%%%%%%%%%%%%%%%%%%%%%%%%%%%%%%%%%%%%%%%%%%%%%%%%%%
% APPENDIX
%%%%%%%%%%%%%%%%%%%%%%%%%%%%%%%%%%%%%%%%%%%%%%%%%%%%%%%%%%%%%%%%%%%%%%%%%%%%%%%
%%%%%%%%%%%%%%%%%%%%%%%%%%%%%%%%%%%%%%%%%%%%%%%%%%%%%%%%%%%%%%%%%%%%%%%%%%%%%%%
\newpage
\appendix
\onecolumn
%\appendix
\clearpage
\section{Appendix}

\subsection{GNN architectures}\label{sec：gnn}
We show results with GraphSage~\citep{hamilton2017representation} architecture as GNN backbone in \tabref{tab:graphsage + neurips split} and \tabref{tab:graphsage + iclr split}. We investigated the two different splits used in previous sections, as well as different self-supervised and supervised pretraining objectives. The overall performance using GraphSage architecture is comparable with results obtained using GIN architecture, and the conclusion about pretraining objectives is the same with what we obtained on the GIN as well. As generally these architectures have similar representation power~\citep{xu2018powerful}, this outcome should be expected. Additionaly, we also explored graph pretraining with graph transformer proposed in \citep{rong2020self}, which is supposed to be more expressive. However, \tabref{tab:grover} in Appendix shows that the results are not competitive. For higher-order GNNs or the deeper GNNs the conclusion might be different, but in general we hold a conservative view towards whether the graph architecture can make a big difference in deciding whether graph pretraining is helpful.

\begin{table*}[h]
\vspace{-3mm}
\centering
\caption{\textbf{GraphSAGE} GNN architecture on Balanced Scaffold Split. The average ROC-AUC without pretraining on all benchmark datasets is 78.1\%. Gray- and transparent- shaded show supervised and unsupervised pretraining objectives. \label{tab:graphsage + neurips split}}
%\vspace{-3mm}
\resizebox{\textwidth}{!}{%
\begin{tabular}{llllllr}
%\Xhline{2\arrayrulewidth}
%\hline
\Xhline{2\arrayrulewidth}
                 & \textbf{BBBP}  & \textbf{BACE}  & \textbf{TOX21} & \textbf{TOXCAST} & \textbf{SIDER} &  \textbf{AVE GAIN}\\ 
%\hline
\Xhline{2\arrayrulewidth}
\nopretrain  & 92.46($\pm$2.91) & 86.46($\pm$1.63) & 78.99($\pm$1.57) & 70.35($\pm$0.09)   & 62.18($\pm$0.77) & 0($\pm$1.394)\\
\masktask     & 93.76($\pm$1.70)  & 88.71($\pm$1.30)  & 78.69($\pm$2.26) & 69.82($\pm$0.36)   & 60.89($\pm$1.67) & 0.286($\pm$1.458)\\
\contexttask & 93.82($\pm$2.35) & 88.61($\pm$1.69) & 79.10($\pm$1.83)  & 70.18($\pm$0.50)    & 62.01($\pm$2.06) & 0.656($\pm$1.686)\\
\Xhline{2\arrayrulewidth}
\rowcolor{Gray}
\supervisedtask  & 93.91($\pm$2.17) & 87.56($\pm$1.86) & 81.06($\pm$1.96) & 71.49($\pm$0.75)   & 63.07($\pm$0.64) & 1.330($\pm$1.476)\\
\rowcolor{Gray}
\supervisedmasktask    & 93.93($\pm$1.50) & 87.62($\pm$1.65) & 80.43($\pm$1.84) & 71.48($\pm$0.86) & 62.93($\pm$0.77) & 1.190($\pm$1.324)\\
\rowcolor{Gray}
\supervisedcontexttask & 92.80($\pm$2.67) & 86.87($\pm$1.89) & 80.62($\pm$1.41) & 71.69($\pm$0.66) & 63.99($\pm$0.52) & 1.106($\pm$1.430)\\ 
%\hline
\Xhline{2\arrayrulewidth}
\end{tabular}%
}
\end{table*}

\begin{table*}[h]
\vspace{-3mm}
\centering
\caption{\textbf{GraphSAGE} GNN architecture on Scaffold Split. The average AUC without pretraining on all 5 datasets is 72.2\%. \label{tab:graphsage + iclr split}}
%\vspace{-3mm}
\resizebox{\textwidth}{!}{%
\begin{tabular}{llllllr}
%\hline
\Xhline{2\arrayrulewidth}

                 & \textbf{BBBP}  & \textbf{BACE}  & \textbf{TOX21} & \textbf{TOXCAST} & \textbf{SIDER}  &  \textbf{AVE GAIN}\\ 
%\hline
\Xhline{2\arrayrulewidth}
\nopretrain  & 74.59($\pm$1.02) & 81.06($\pm$0.15) & 75.72($\pm$0.34) & 66.44($\pm$0.27)   & 63.31($\pm$0.57)  & 0($\pm$0.470)\\
\masktask    & 75.18($\pm$0.52) & 82.98($\pm$0.62) & 75.50($\pm$0.33)  & 67.32($\pm$0.21)   & 64.26($\pm$0.06)  & 0.824($\pm$0.348)\\
\contexttask & 74.89($\pm$0.47) & 82.19($\pm$0.59) & 75.45($\pm$0.24) & 67.14($\pm$0.02)   & 64.22($\pm$0.06)  & 0.554($\pm$0.276)\\
\Xhline{2\arrayrulewidth}
\rowcolor{Gray}
\supervisedtask  & 74.58($\pm$0.44) & 81.56($\pm$0.77) & 76.65($\pm$0.22) & 68.26($\pm$0.17)   & 63.50($\pm$0.25)   & 0.686($\pm$0.370)\\
\rowcolor{Gray}
\supervisedmasktask     & 75.66($\pm$0.45) & 84.14($\pm$0.37) & 76.92($\pm$0.08) & 67.96($\pm$0.20)  & 64.70($\pm$0.12)   & 1.652($\pm$0.244)\\
\rowcolor{Gray}
\supervisedcontexttask & 76.41($\pm$0.19) & 81.59($\pm$0.49) & 77.61($\pm$0.17) & 67.92($\pm$0.19) & 64.92($\pm$0.15)  & 1.466($\pm$0.238)\\ 
%\hline
\Xhline{2\arrayrulewidth}
\vspace{-8mm}
\end{tabular}%
}
 % \vspace{-100pt}
\end{table*}

\subsection{Graph Parameters: GNN layers}
\label{sec:graph_layers}
%\textbf{Graph Parameters}
We also investigate the effect of the number of GNN layers in Graph pretraining. We use the recommended graph parameters in \cite{hu2019strategies} (number of GNN layer=5), which are selected for best performance.  Table \ref{GNNlayer  neuripssplit} below is using node mask pretraining for GNN layer = 7. Results of GNN layer = 5 is presented in Table 2 and 4. Compared with layer 5 vs 7, the conclusion that ``pretraining does not help statistically significantly" does not change. 
\begin{table}[hb]
%\small
%\centering
%\vspace{-6mm}
\caption{\textbf{Tune number of GNN layers}: Self-supervised + Rich feature.  Layer number=7 \label{GNNlayer  neuripssplit}}
\resizebox{\textwidth}{!}{%
\begin{tabular}{l|l|ccccc}
\hline
\multicolumn{1}{l|}{\textbf{GNN layers}} & \textbf{Methods}       & \textbf{BBBP}   & \textbf{BACE}   & \textbf{TOX21} & \textbf{TOXCAST} & \textbf{SIDER}                       \\ \hline
%\multirow{2}{*}{\textbf{5}}                       & \nopretrain  & 92.23 ($\pm$3.07) & 87.43 ($\pm$1.63) & 79.20 ($\pm$1.99) & 69.13 ($\pm$0.55)  & \multicolumn{1}{r}{61.92($\pm$0.89)} \\
%                   & \masktask  & 92.24 ($\pm$2.76)         & 87.32($\pm$1.67) & 79.57 ($\pm$2.03) & 69.77 ($\pm$0.13) & \multicolumn{1}{r}{61.62($\pm$1.12)} \\ \hline
\multirow{2}{*}{\textbf{Balanced Scaffold Split}} & \nopretrain   &  92.53($\pm$1.96) & 87.01($\pm$1.28) & 78.97($\pm$1.78)  & 69.07($\pm$0.45)  & 61.35($\pm$1.51)                       \\
                   & \masktask         & 92.68($\pm$2.33)          & 86.98($\pm$2.71) & 79.20($\pm$1.97)   & 69.83($\pm$0.73)  & 61.59($\pm$1.05)                       \\ %\hline
 \Xhline{2\arrayrulewidth}
 \multirow{2}{*}{\textbf{Scaffold Split}} & \nopretrain  &  74.64($\pm$1.28) & 79.85($\pm$0.02)   & 75.75($\pm$0.79)    & 66.09($\pm$0.29)   & 61.81($\pm$0.41)       \\
                   & \masktask    & 74.94($\pm$1.01)          & 81.33($\pm$1.12)   & 75.81($\pm$0.46)    & 66.39($\pm$0.23)   & 63.80($\pm$0.18)   \\      % \hline
  \Xhline{2\arrayrulewidth}
  %\vspace{-2 mm}
\end{tabular}%
 %\vspace{-25 mm}
}
%\caption{GNNlayer + neuripssplit}
%\label{tab:GNNlayer  neuripssplit}
\end{table}

\subsection{Selection of Dataset}
 As many pretraining objectives to be evaluated are following \cite{hu2019strategies, rong2020self}, we used the intersection dataset of the two papers, except we eliminate CLINTOX. Because CLINTOX has a significant high variance in balanced scaffold split even without pretraining (see Table \ref{tab:clintox} and \ref{tab:clintox-all} below), so we remove it from the evaluation to avoid unstable and biased evaluation. Additionally, we provide CLINTOX results (Table \ref{tab:clintox} and \ref{tab:clintox-all}). The conclusion is the same on CLINTOX as on the other datasets.
 
% \begin{wraptable}{r}{0.5\linewidth}
%\small
%\vspace{-12mm}
%\caption{\textbf{Clintox}:High variance on \textbf{Balanced Scaffold Split}. \label{clintox}}
%\centering
\begin{table}
\centering
\caption{\textbf{Clintox}:High variance on \textbf{Balanced Scaffold Split}.}\label{tab:clintox}
%\caption{\textbf{Clintox}:High variance on \textbf{Balanced Scaffold Split}. }\label{clintox}
\begin{tabular}{lll}
\hline
\textbf{No Pretrain}             & \textbf{Layer = 5} & \textbf{Layer = 7} \\ \hline
{Balanced Scaffold Split} & 77.47({\color{red}$\pm$10.44})    & 77.45({\color{red}$\pm$9.26})     \\
{Scaffold Split}          & 89.70($\pm$0.93)     & 88.50($\pm$0.85)     \\ \hline
\end{tabular}%
%\end{wraptable} 
\end{table}

% \usepackage{graphicx}
\begin{table}[ht]
\caption{Self-supervised + rich feature + GNN layer = 7}\label{tab:clintox-all}
\resizebox{\textwidth}{!}{%
\begin{tabular}{l|l|llllll}
\hline
                        & \textbf{Methods}         & \textbf{BBBP}           & \textbf{BACE }          & \textbf{CLINTOX }       & \textbf{TOX21}          & \textbf{TOXCAST}        & \textbf{SIDER }         \\ \hline
\textbf{Balanced Scaffold Split} & No pretrain     & 92.53(+/-1.96) & 87.01(+/-1.28) & 77.45(+/-9.26) & 78.97(+/-1.78) & 69.07(+/-0.45) & 61.35(+/-1.51) \\
                        & Node prediction & 92.68(+/-2.33) & 86.98(+/-2.71) & 79.63(+/-9.21) & 79.2(+/-1.97)  & 69.83(+/-0.73) & 61.59(+/-1.05) \\ \hline
\textbf{Scaffold Split }         & No pretrain     & 74.64(+/-1.28) & 79.85(+/-0.02) & 88.5(+/-0.85)  & 75.75(+/-0.79) & 66.09(+/-0.29) & 61.81(+/-0.41) \\
                        & Node predicion  & 74.94(+/-1.01) & 81.33(+/-1.12) & 87.16(+/-0.99) & 75.81(+/-0.46) & 66.39(+/-0.23) & 63.8(+/-0.18)  \\ \hline
\end{tabular}%
}
\end{table}

\subsection{Reproduce existing results}
\label{sec:reproduce}

\begin{table*}[ht]
\centering
\caption{Test ROC-AUC (\%) performance on molecular property benchmarks using {\it unsupervised} and {\it supervised} pre-training objectives (self-supervised). Unlike the basic framework, the results are generated using {\it Basic} features (not rich feature), scaffold split without averaging over three different data splits, and no selection of the best performance on six learning rates. All these factors lead to conclusion in favor of pretraining. }
\label{tab:feature-balancedscaffold-basicfeature-wrong}
% \resizebox{\textwidth}{!}{%
\begin{tabular}{lccccc}
%\hline
\Xhline{2\arrayrulewidth}
\textbf{ICLR code} & \textbf{BBBP} & \textbf{BACE} & \textbf{TOX21} & \textbf{TOXCAST} & \textbf{SIDER} \\ 
%\hline
\Xhline{2\arrayrulewidth}
\nopretrain             & 89.82 & 79.46 & 77.39 & 67.91 & 58.38 \\
\masktask                 & 90.42 & 84.03 & 78.70 & 68.36 & 59.97 \\
\contexttask             & 91.07 & 83.41 & 78.84 & 68.57 & 60.78 \\ \Xhline{1.5\arrayrulewidth}
\rowcolor{Gray}
\supervisedtask             & 89.37 & 82.74 & 79.78 & 68.39 & 63.30 \\
\rowcolor{Gray}
\supervisedmasktask    & 91.53 & 83.94 & 81.34 & 70.93 & 62.46 \\
\rowcolor{Gray}
\supervisedcontexttask & 91.00 & 82.90 & 81.23 & 71.34 & 62.72 \\ 
%\hline
\Xhline{2\arrayrulewidth}
\end{tabular}%
% }
%\caption{Balanced scaffold + basic feature}

\end{table*}

We first reprudce the results from~\citet{hu2019strategies} in \tabref{tab:feature-balancedscaffold-basicfeature-wrong}, where we make sure the code base can obtain the similar results where the pretraining seems to be helpful. However as in our analysis, The three factors (feature engineering, data splits, tuning of hyper-parameters) all contribute positively to the results, and the effect of these factors could be even larger than the pretraining itself.

% \usepackage{graphicx}
% \usepackage{graphicx}
\begin{table*}[ht]
\caption{Reproduce Grover \citep{rong2020self} \label{tab:grover}}
\resizebox{\textwidth}{!}{%
\begin{tabular}{lllllll}
\hline
 \textbf{GROVER }       & \textbf{Number of parameters} & \textbf{BBBP}  & \textbf{BACE}  & \textbf{TOX21} & \textbf{TOXCAST} & \textbf{SIDER} \\ \hline
No pretrain (base)  & 48,790,038  & 92.6 (+/- 2.0) & 86.2 (+/- 2.4) & 79.3 (+/- 2.6) & 71.5 (+/- 0.3) & 62.3 (+/- 0.3) \\
No pretrain (large) & 107,714,488 & 91.8 (+/- 2.9) & 86.4 (+/- 2.1)   & 79.8 (+/- 2.2) & 71.5 (+/- 4.0) & 63.6 (+/- 0.7) \\
grover pretraining (base)  & 48,790,038                    & 89.6 (+/- 0.8) & 85.1 (+/- 1.1) & 78.7 (+/- 2.4) & 71.5 (+/- 0.5)   & 62.0 (+/- 2.4) \\
grover pretraining (large) & 107,714,488                   & 89.2 (+/- 0.7) & 84.6 (+/- 1.4) & 78.6 (+/- 2.6) & 69.1 (+/- 2.4)   & 61.4 (+/- 1.9) \\ \hline
\end{tabular}%
 }
\end{table*}

\tabref{tab:grover} shows the results we reproduced for GROVER \citep{rong2020self}. We compared the fine-tuning results with the model initialized from the pretrained ones provided in their website. However we didn't see significant gains over the model trained from scratch. We believe the hyper-parameters matters more in this case. %Unfortunately we didn't receive the reply from the authors of GROVER about our inquery. Nevertheless, we present our empirical findings here for reference.

\subsection{Molecular benchmark description} \label{sec:molecular_benchmarks}
\begin{itemize}
    \item \textbf{BBBP}: \\
    The Blood-brain barrier penetration (BBBP) dataset is extracted from a study on the modeling and prediction of the barrier permeability. As a membrane separating circulating blood and brain extracellular fluid, the blood-brain barrier blocks most drugs, hormones and neurotransmitters. Thus penetration of the barrier forms a long-standing issue in development of drugs targeting central nervous system. This dataset includes binary labels for over 2000 compounds on their permeability properties.
    References:
Martins, Ines Filipa, et al. "A Bayesian approach to in silico blood-brain barrier penetration modeling." Journal of chemical information and modeling 52.6 (2012): 1686-1697.
~     

    \item \textbf{BACE} \\
    The BACE dataset provides quantitative (IC50) and qualitative (binary label) binding results for a set of inhibitors of human $\beta$-secretase 1 (BACE-1). All data are experimental values reported in scientific literature over the past decade, some with detailed crystal structures available. A collection of 1522 compounds with their 2D structures and properties are provided.

References:
Subramanian, Govindan, et al. "Computational modeling of $\beta$-secretase 1 (BACE-1) inhibitors using ligand based approaches." Journal of chemical information and modeling 56.10 (2016): 1936-1949.
    \item \textbf{TOX21}
    
    The “Toxicology in the 21st Century” (Tox21) initiative created a public database measuring toxicity of compounds, which has been used in the 2014 Tox21 Data Challenge. This dataset contains qualitative toxicity measurements for 8k compounds on 12 different targets, including nuclear receptors and stress response pathways.

The data file contains a csv table, in which columns below are used:
     "smiles" - SMILES representation of the molecular structure
     "NR-XXX" - Nuclear receptor signaling bioassays results
     "SR-XXX" - Stress response bioassays results
        please refer to the links at \url{https://tripod.nih.gov/tox21/challenge/data.jsp} for details.

References:
Tox21 Challenge. \url{https://tripod.nih.gov/tox21/challenge/}
    \item \textbf{TOXCAST}
    
    ToxCast is an extended data collection from the same initiative as Tox21, providing toxicology data for a large library of compounds based on in vitro high-throughput screening. The processed collection includes qualitative results of over 600 experiments on 8k compounds.

The data file contains a csv table, in which columns below are used:
     "smiles" - SMILES representation of the molecular structure
     "ACEA\_T47D\_80hr\_Negative", \\ "Tanguay\_ZF\_120hpf\_YSE\_up" - Bioassays results
        please refer to the section "high-throughput assay information" at \url{https://www.epa.gov/chemical-research/toxicity-forecaster-toxcasttm-data} for details.

References:
Richard, Ann M., et al. "ToxCast chemical landscape: paving the road to 21st century toxicology." Chemical research in toxicology 29.8 (2016): 1225-1251.

    \item \textbf{SIDER}
    
    The Side Effect Resource (SIDER) is a database of marketed drugs and adverse drug reactions (ADR). The version of the SIDER dataset in DeepChem has grouped drug side effects into 27 system organ classes following MedDRA classifications measured for 1427 approved drugs.

The data file contains a csv table, in which columns below are used:
     "smiles" - SMILES representation of the molecular structure
     "Hepatobiliary disorders" ~ "Injury, poisoning and procedural complications" - Recorded side effects for the drug
        Please refer to \url{http://sideeffects.embl.de/se/?page=98} for details on ADRs.

References:
Kuhn, Michael, et al. "The SIDER database of drugs and side effects." Nucleic acids research 44.D1 (2015): D1075-D1079.
Altae-Tran, Han, et al. "Low data drug discovery with one-shot learning." ACS central science 3.4 (2017): 283-293.
Medical Dictionary for Regulatory Activities. \url{http://www.meddra.org/}

\end{itemize}

\clearpage
%%%%%%%%%%%%%%%%%%%%%%%%%%%%%%%%%%%%%%%%%%%%%%%%%%%%%%%%%%%%

%%%%%%%%%%%%%%%%%%%%%%%%%%%%%%%%%%%%%%%%%%%%%%%%%%%%%%%%%%%%

\end{document}